\documentclass{article}

\PassOptionsToPackage{square,sort,comma,numbers}{natbib}

\usepackage[final]{neurips_2021}

\usepackage[utf8]{inputenc} %
\usepackage[T1]{fontenc}    %
\usepackage{hyperref}       %
\usepackage{url}            %
\usepackage{booktabs}       %
\usepackage{amsfonts}       %
\usepackage{nicefrac}       %
\usepackage{microtype}      %
\usepackage{xcolor}         %

\usepackage{mathtools}
\usepackage{dsfont}
\usepackage{cleveref}
\usepackage{subcaption}
\usepackage{wrapfig}
\usepackage{color}
\usepackage{amsmath}
\usepackage{amssymb}
\usepackage{multirow, varwidth}
\usepackage{dblfloatfix}
\usepackage{verbatim}
\usepackage{xspace}
\makeatletter
\newif\if@restonecol
\makeatother

\usepackage{xr}
\usepackage[amsmath,thmmarks]{ntheorem}

\DeclareRobustCommand\onedot{\futurelet\@let@token\@onedot}
\def\onedot{.\xspace}
\def\eg{\emph{e.g}\onedot} 
\def\ie{\emph{i.e}\onedot}

\definecolor{candypink}{rgb}{0.89, 0.44, 0.48}
\definecolor{mediumaquamarine}{rgb}{0.4, 0.8, 0.67}
\definecolor{azure}{rgb}{0.0, 0.5, 1.0}
\definecolor{awesome}{rgb}{1.0, 0.13, 0.32}

\newcommand{\dt}{\delta}
\newcommand{\tmax}{\text{max}}
\newcommand{\du}{t}
\newcommand{\sr}{d}

\newtheorem{theorem}{Theorem}
\newtheorem{proposition}[theorem]{Proposition}

\title{Time Discretization-Invariant\\Safe Action Repetition for Policy Gradient Methods}

\author{
  Seohong Park\\
  Seoul National University\\
  \texttt{artberryx@snu.ac.kr} \\
  \And
  Jaekyeom Kim\\
  Seoul National University\\
  \texttt{jaekyeom@snu.ac.kr} \\
  \And
  Gunhee Kim\\
  Seoul National University\\
  \texttt{gunhee@snu.ac.kr} \\
}

\begin{document}

\maketitle

\begin{abstract}
    In reinforcement learning, continuous time is often discretized by a time scale $\dt$,
    to which the resulting performance is known to be highly sensitive.
    In this work, we seek to find a \textit{$\dt$-invariant} algorithm for policy gradient (PG) methods,
    which performs well regardless of the value of $\dt$.
    We first identify the underlying reasons that cause PG methods to fail as $\dt \to 0$,
    proving that the variance of the PG estimator can diverge to infinity in stochastic environments
    under a certain assumption of stochasticity.
    While durative actions or action repetition can be employed to have $\dt$-invariance,
    previous action repetition methods cannot immediately react to unexpected situations in stochastic environments.
    We thus propose a novel $\dt$-invariant method named \textit{Safe Action Repetition (SAR)}
    applicable to any existing PG algorithm.
    SAR can handle the stochasticity of environments by adaptively reacting to changes in states during action repetition.
    We empirically show that our method is not only $\dt$-invariant but also robust to stochasticity,
    outperforming previous $\dt$-invariant approaches on eight MuJoCo environments with
    both deterministic and stochastic settings.
    Our code is available at \url{https://vision.snu.ac.kr/projects/sar}.
\end{abstract}

\section{Introduction}
\label{sec:intro}

Deep reinforcement learning (RL) has demonstrated phenomenal achievements in a wide array of tasks, including superhuman game-playing \citep{atari_mnih2015,muzero_schrittwieser2020} and controlling complex robots \citep{gu2017,qtopt_salashnikov2018}.
Most RL algorithms are based on an Markov Decision Process (MDP), which is a discrete-time control process for the iteration of observing a state and performing an action.
However, numerous real-world problems such as robotic manipulation and autonomous driving are defined in continuous time, which does not directly fit the MDP setting.
To fill this gap, continuous time is often discretized by a \textit{discretization time scale} $\dt$, where the RL agent makes a decision at every $\dt$.
It has been shown that RL algorithms are greatly sensitive to this hyperparameter \citep{dau_tallec2019,what_andrychowicz2021}.
For instance, altering $\dt$ via frame skipping leads to drastic performance differences \citep{frameskip_braylan2015,what_andrychowicz2021}.
Indeed, an excessively high $\dt$ precludes the agent from making fine-grained decisions, which is likely to cause performance degradation.

On average, the agent could perform equally well or better with a lower $\dt$ than with a higher $\dt$, since the agent can make decisions more frequently.
However, \citet{baird1994} and \citet{dau_tallec2019} theoretically proved that the standard Q-learning fails when $\dt \to 0$ as the action-value (Q) function collapses to the state-value (V) function, eliminating preferences between actions.
As will be shown in \Cref{sec:dt0}, policy gradient (PG) methods fail as well with an infinitesimal $\dt$ for the following three reasons: 
(1) The variance of the gradient estimator explodes. %
(2) Exploration ranges may become highly limited.
(3) Infinitely many decision steps are required.
The latter two also apply to Q-learning methods.

Therefore,
it is generally required to differently set an appropriate $\dt$ for each continuous environment.
Indeed, continuous control environments in MuJoCo \citep{mujoco_todorov2012} have different discretization time scales from one another, ranging from 0.008s (Hopper) to 0.05s (InvertedDoublePendulum).
However, such tuning of $\dt$ could be burdensome when applying RL algorithms to new environments,
considering its significant influence on performance \citep{what_andrychowicz2021,dau_tallec2019}.
Furthermore, even if the optimal $\dt$ is found in simulation, trained policies may not be transferable to real-world settings, since physical sensors often have their unique sampling frequencies. 

There have been proposed some methods for robustness to discretization of time scales.
$\dt$-invariant methods could bring several advantages:
(1) It obviates the need for tuning $\dt$ on each continuous control environment.
(2) They can achieve better performance
by utilizing more fine-grained control with a low $\dt$.
(3) Using a policy with adaptive decision frequencies (as a variant of $\dt$-invariant policies),
the agent can efficiently take actions only when necessary, which could expedite training without losing agility.
\citet{dau_tallec2019} introduced an algorithm based on Advantage Updating \citep{baird1994},
which can make existing Q-learning methods (\eg, DQN \citep{dqn_mnih2013} and DDPG \citep{ddpg_lillicrap2016})
invariant to $\dt$ by preventing them from Q-function collapse.
For PG methods, %
\citet{ct_munos2005} and \citet{wawrzynski2015} proposed methods that can cope with fine time discretization.
However, these methods either assume access to the gradient of the reward function or require an infinite number of decision steps (or training steps) when $\dt \to 0$, both of which could hinder its application to real-world environments.

We aim at proposing an efficient $\dt$-invariant approach applicable to existing PG methods such as PPO \citep{ppo_schulman2017}, TRPO \citep{trpo_schulman2015} and A2C \citep{a2c_mnih2016}.
One straightforward approach may be to take \textit{durative actions} by making policies produce both \textit{actions} and their \textit{durations}.
Such an approach is practically equivalent to prior work on \textit{action repetition} whose policies output both actions and the number of action repetitions \citep{dfdqn_lakshminarayanan2017,figar_sharma2017},
since continuous control environments such as MuJoCo often already provide discretized time scales.
However, prior approaches to action repetition possess some limitations.
For example, there is no way to stop repeating a chosen action during a repetition period,
which means that they are not capable of immediately handling unexpected events in stochastic environments.
This could lead to catastrophic failure in some real-world settings such as autonomous driving.

We thus propose an alternative approach named  \textit{Safe Action Repetition} (SAR) with the key idea of repeating an action until the agent exits its \textit{safe region}.
Our policy produces both an \textit{action} and a \textit{safe region} in the state space, only within which the chosen action is repeated.
SAR enables any PG algorithm to not only be $\dt$-invariant but also be robust to stochasticity such as unexpected events in the environment,
because such situations lead the agent's state to be outside of the safe region, immediately causing the cease of the current action.
We apply the proposed method to several PG algorithms and empirically show that SAR indeed exhibits $\dt$-invariance on various MuJoCo environments and outperforms baselines on both deterministic and stochastic settings.

Our contributions can be summarized as follows:
\begin{itemize}
    \item We first provide a more general proof on the variance explosion of the PG estimator, which is the main reason why PG methods fail as $\dt \to 0$.
        We then show that temporally extended actions can resolve the failure mode of PG algorithms with a low $\dt$.
    \item We introduce a novel $\dt$-invariant method named SAR applicable to any PG method on continuous control domains.
        To the best of our knowledge, this is the first action repetition (or durative action) method that repeats an action based on the agent's \textit{state}, rather than a precomputed action duration.
        As a result, SAR can cope with unexpected situations in stochastic environments, which existing action repetition methods cannot handle.
    \item We apply SAR to three PG methods, PPO, TRPO and A2C, and empirically demonstrate that our SAR method is mostly invariant to $\dt$ on eight MuJoCo environments.
        We also verify its robustness to stochasticity via three different stochastic settings on each MuJoCo environment.
        Our method also outperforms previous $\dt$-invariant approaches such as FiGAR-C \citep{figar_sharma2017} and DAU \cite{dau_tallec2019} on those settings.
\end{itemize}

\section{Related Work}
\label{sec:related}

\textbf{Continuous-time RL.}
Reinforcement learning in continuous-time domains has long been studied with various approaches
\cite{ct_munos1996,ct_bradtke1994,ct_doya2000,ct_munos2005,ode_du2020,baird1994,ct_frmaux2013}.
\citet{ct_bradtke1994} extended existing Q-learning and temporal difference methods to semi-MDPs, which can be viewed as a continuous-time generalization of MDPs.
\citet{ct_doya2000} developed a continuous actor-critic method based on the Hamilton-Jacobi-Bellman (HJB) equation, a continuous-time counterpart of the Bellman equation, approximating policies and value functions with radial basis functions.

\textbf{Time discretization.}
Another line of research to cope with continuous-time environments is to use finely discretized MDPs. %
\citet{baird1994} first informally presented that when $\dt \to 0$, two different Q values on the same state will eventually collapse to the V value, such that $Q(s, a_1) \approx Q(s, a_2) \approx V(s)$, since the influence of each action is vanished by the infinitesimal time scale.
They proposed \textit{Advantage Updating} as a solution to avoid the collapse by appropriately scaling the advantage function $A(s, a) = \frac{Q(s, a) - V(s)}{\dt}$.
More recently, \citet{dau_tallec2019} theoretically proved the existence of collapse in low-$\dt$ settings and extended Advantage Updating for deep neural networks, showing its $\dt$-invariance on classic control benchmarks.
For PG methods, \citet{ct_munos2005} first demonstrated that the variance of the policy gradient estimate can be infinite when $\dt \rightarrow 0$,
and proposed an algorithm based on pathwise derivatives, in which the variance of the estimator decreases to 0 when $\dt \to 0$.
However, it assumes that the gradient of the reward function $\nabla r(s, a)$ is known to the agent.
\citet{wawrzynski2015,ar_korenkevych2019} proposed methods based on autocorrelated noise that could prevent the variance explosion.
Notably, all of these approaches choose an action at every $\dt$. However, this makes it infeasible to train when $\dt$ is nearly zero as the number of decision steps goes to infinity.
In our work, we employ durative actions to achieve $\dt$-invariance on PG methods,
which could resolve the problem of variance explosion and infinite decision steps.

\textbf{Action repetition.}
Our proposed method is closely related to existing action repetition methods.
Most works on action repetition let the policy also determine how many times (or how long) an action is repeated.
DFDQN \citep{dfdqn_lakshminarayanan2017} doubled the action space by mapping a half of it to actions repeated $r_1$ times and the other half to those repeated $r_2$ times, where $r_1$ and $r_2$ are fixed hyperparameters.
FiGAR \citep{figar_sharma2017} introduced a repetition policy $\pi(x|s)$ in addition to the original action policy $\pi(a|s)$ so that the agent repeats the action $x$ times,
where $x$ is selected from a predefined set $W$; \eg, $W = \{1, 2, \ldots, 30\}$.
\citet{pfqi_metelli2020} theoretically analyzed the performance of the optimal policy when a fixed action repetition count is given, and proposed a heuristic to approximately choose the optimal control frequency.
On the other hand, we take a completely different approach where our policy produces not repetition counts (or action durations) but \textit{safe regions},
which enables the agent to adaptively stop action repetition when facing unexpected events in stochastic environments.

Finally, action repetition is related to the options framework \citep{hrl_sutton1999}
in that they both use temporally extended actions.
In the options framework, the agent learns both a high-level inter-option policy and a low-level intra-option policy,
where action repetition can be interpreted as a special case of an intra-option policy.
However, one crucial difference between them is that
within the options framework, the agent has to produce each $\dt$-discretized low-level action (even with open-loop options),
which makes having $\dt$-invariance non-trivial,
unlike the action repetition approach.

\section{Preliminaries}

We consider a continuous-time MDP $\mathcal{M} = (\mathcal{S}, \mathcal{A}, r, F, \gamma)$ \citep{ct_bradtke1994,ct_doya2000},
where $\mathcal{S}$ is a continuous state space, $\mathcal{A}$ is a bounded action space, $F \colon \mathcal{S} \times \mathcal{A} \to \mathcal{S}$ is a transition dynamics function,
$r \colon \mathcal{S} \times \mathcal{A} \to \mathbb{R}$ is a reward function and $\gamma \in (0, 1]$ is a discount factor.
For simplicity, we assume the environment has deterministic transition dynamics;
we refer to \citet{ct_munos1996} for the stochastic case involving Brownian motion.
The transition dynamics and the return $R(\tau)$ are given by
\begin{align}
  s(t) = s(0) + \int_0^t F(s(t'), a(t')) dt', \hspace{12pt}
    R(\tau) = \int_0^{\infty} \gamma^{t'} r(s(t'), a(t')) dt', 
\end{align}
where $s(t)$ and $a(t)$ respectively denote the state and action at time $t$, and $\tau$ denotes the whole trajectory consisting of states and actions.

Following \citet{dau_tallec2019}, we define a discretized version of $\mathcal{M}$ as $\mathcal{M}_{\dt} = (\mathcal{S}, \mathcal{A}, r_{\dt}, F_{\dt}, \gamma_{\dt})$ with a \textit{discretization time scale} $\dt > 0$, where the agent observes a state and performs an action at every $\dt$.
We respectively denote the state and action at $i$-th step as $s_i$ and $a_i$, where $s_i$ in $\mathcal{M}_{\dt}$ corresponds to $s(i \dt)$ in $\mathcal{M}$ and $a_i$ is maintained during the time interval $[i\dt, (i+1)\dt)$.
In $s(t)$, $t$ indicates the \textit{physical} time, and $i$ in $s_i$ is the number of steps taken.
The reward $r_i$ and return $R_{\dt}(\tau)$ are defined as 
\begin{align}
    r_i = r_{\dt}(s_i, a_i) = r(s(i\dt), a(i\dt)) \dt, \hspace{12pt}
    R_{\dt}(\tau) = \sum_{i=0}^{\infty} \gamma_{\dt}^i r_i \label{eq:delta_reward},
\end{align}
where the discount factor is $\gamma_{\dt}= \gamma^{\dt}$. 
Given a deterministic policy $\pi \colon \mathcal{S} \to \mathcal{A}$,
the continuous value function $V^\pi(s)$ and the discretized one $V_{\dt}^\pi$ are defined as
$V^\pi(s) = \mathbb{E}_{\tau \sim p_\pi(\tau)} [R(\tau) | s(0) = s]$ and
$V_{\dt}^\pi(s) = \mathbb{E}_{\tau \sim p_\pi(\tau)} [R_{\dt}(\tau) | s_0 = s]$, respectively.
\citet{dau_tallec2019} proved that $V_{\dt}^\pi$ converges to $V^\pi$ when $\dt \to 0$ under smoothness assumptions.

In the rest of the paper, we will focus on $\mathcal{M}_{\dt}$ (\ie, $\mathcal{M}$ with time discretization) and omit the subscript $\dt$ unless it is necessary.
Also, as in ordinary discrete-time MDPs \citep{rl_sutton2018}, we consider stochastic transition dynamics $p(s_{i+1} | s_i, a_i)$ and stochastic policy $\pi(a_i | s_i)$ instead of deterministic ones.

\section{Safe Action Repetition}
\label{sec:method}

We first show that $\dt \to 0$ leads to failure in policy gradient (PG) methods (\Cref{sec:dt0}),
and propose that durative actions or action repetition can be a solution to the failure mode.
We then point out that existing action repetition methods have drawbacks in the presence of unexpected events in stochastic environments (\Cref{sec:durative}).
As a solution, we propose \textit{Safe Action Repetition} (SAR) as a novel $\dt$-invariant approach for PG algorithms, which is robust to such stochasticity (\Cref{sec:sar}).
We additionally suggest a variant of SAR to deal with non-Markovian environments (\Cref{sec:sar_nme}).

\subsection{Policy Gradients with Infinitesimal Discretization of Time Scale}
\label{sec:dt0}

A smaller discretization time scale should not lead to worse maximum returns on average, since it allows the agent to perform more fine-grained actions.
However, both Q-learning and PG methods are subjected to fail with a too small $\dt$.
We introduce three reasons why PG methods fail. We refer to \citet{baird1994,dau_tallec2019} for further discussion on Q-learning methods.

\textbf{Variance explosion of the policy gradient estimator.}
We show that the variance of the PG estimator can diverge to infinity with a decrease in $\dt$.
While this was first shown in \citet{ct_munos2005} via a simple illustrative example,
we provide a more general proof without assuming a particular environment.
Specifically, we prove that the variance explosion problem can arise in any stochastic environment
where the variance of its return conditioned on actions is greater than a small positive constant.

Let us consider a stochastic environment that has a finite physical time limit $T$ and a discretization time scale $\dt$.
It follows that the number of decision steps (or actions) in a single rollout is $N = T/\dt$.
Let the policy $\pi_\theta(a_i|s_i)$ be parameterized by $\theta$, the distribution over trajectories $\tau = (s_0, a_0, \ldots, s_N)$
be given by $p_\theta(\tau) = p(s_0)\prod_{i=0}^{N-1} \pi_\theta(a_i|s_i)p(s_{i+1}|s_i,a_i)$ and $p_\theta(s_{0:N})$ denote its state-marginal distribution.
For simplicity, we assume that
the policy is represented as a multivariate normal distribution with a learnable diagonal covariance matrix:
$\pi_\theta(a_i|s_i) \sim \mathcal{N}(\mu_{\theta_\mu}(s_i), \Sigma)$,
where the mean $\mu_{\theta_\mu}(s_i) = [\mu_{\theta_\mu,1}(s_i), \ldots, \mu_{\theta_\mu,K}(s_i)]^\top$
is modeled by a neural network,
$\Sigma = \mathrm{diag}(\sigma^2_1, \ldots, \sigma^2_K)$ is the learnable variance that is independent of states
(as in the original TRPO \cite{trpo_schulman2015} and PPO \cite{ppo_schulman2017}).
Thus, $\theta = [\theta_\mu^\top, \sigma_1, \ldots, \sigma_K]^\top$ is the whole parameters of the policy $\pi_\theta$,
and $K = \mathrm{dim}(\mathcal{A})$.

The derivative of the RL objective function $J(\theta) = \mathbb{E}_{\tau \sim p_\theta(\tau)} [R(\tau)]$ can be written as 
\begin{align}
    \nabla_\theta J(\theta) &= \mathbb{E}_{\tau \sim p_\theta(\tau)} \left[ \left(\sum_{i=0}^{N-1} \nabla_\theta \log \pi_\theta (a_i | s_i)\right) R(\tau) \right]
    \triangleq \mathbb{E}_{\tau \sim p_\theta(\tau)} [G_\theta(\tau)] \label{eq:g_tau},
\end{align}
which is often referred to as the \textit{policy gradient estimator}.

We derive a lower bound for its total variation $\mathrm{tr} [ \mathbb{V}_{\tau \sim p_\theta(\tau)} [ G_\theta(\tau) ] ]$,
where $\mathbb{V}[X]$ is the variance of a variable $X$ (or the covariance matrix when $X$ is multidimensional), and $\mathrm{tr}$ is the trace operator.

\begin{theorem}
    \label{th:variance}
    If the environment is stochastic in the sense that
    for any reparameterized actions $\epsilon_{0:N-1}$,
    if the variance of returns conditioned on the actions is lower bounded by a small positive constant $c$
    (\ie, $\mathbb{V}_{s_{0:N} \sim p_\theta(s_{0:N}|\epsilon_{0:N-1})}\left[ R(\tau) \right] \geq c > 0$,
    where $a_i = \mu_{\theta_\mu}(s_i) + \Sigma \epsilon_i$ and $\epsilon_i \overset{\text{i.i.d.}}{\sim} \mathcal{N}(0, I)$), 
    it holds that
    \begin{align}
        \mathrm{tr} \left[ \mathbb{V}_{\tau \sim p_\theta(\tau)} \left[ G_\theta(\tau) \right] \right]
        \geq \frac{Tc}{\delta \cdot \mathrm{min}(\sigma_1^2, \sigma_2^2, \ldots, \sigma_K^2)} \label{eq:pgexplosion}.
    \end{align}
\end{theorem}

We provide a proof and further discussion in \Cref{sec:proof1}.
\Cref{eq:pgexplosion} indicates that $\dt \to 0$ can lead the variance of the PG estimator to explode, especially in stochastic environments.
We emphasize that the two primary causes of this explosion are the independence of actions and the infinitely growing number of decision steps,
both of which correspond to \Cref{eq:cause_explosion1} $\to$ (\ref{eq:cause_explosion2}) in \Cref{sec:proof1}.
Practically, existing PG algorithms are implemented with
standard variance reduction techniques such as
reward-to-go policy gradient and baseline functions. %
However,
even if such techniques are applied,
the variance of the PG estimator is still likely to explode as $\dt \to 0$ considering the environment's stochasticity,
if the learning rate and minibatch size remain the same.

\begin{wrapfigure}{r}{0.26\textwidth}
  \centering
  \raisebox{0pt}[\dimexpr\height-1.2\baselineskip\relax]{
    \begin{subfigure}[t]{0.49\linewidth}
      \includegraphics[width=\linewidth]{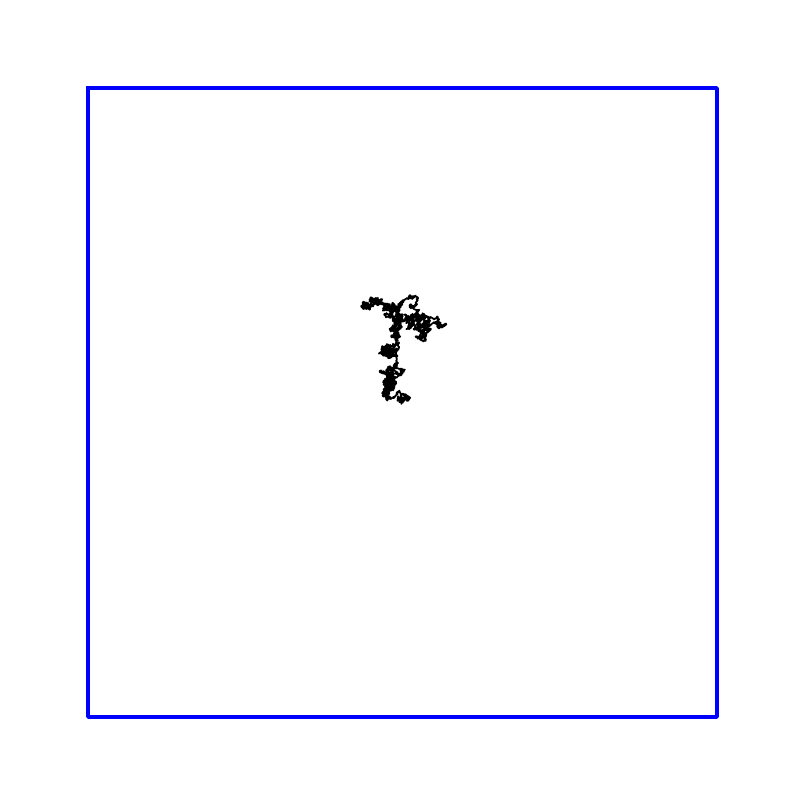}
      \caption{Low $\dt$}
    \end{subfigure}
    \begin{subfigure}[t]{0.49\linewidth}
      \includegraphics[width=\linewidth]{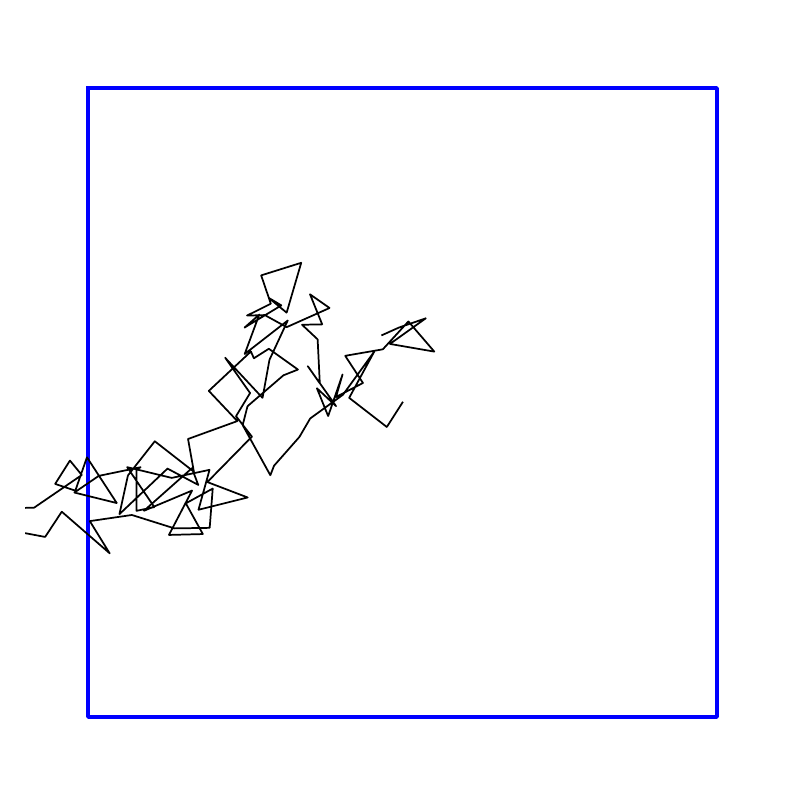}
      \caption{High $\dt$}
    \end{subfigure}
  }
  \caption{
      An example of 2-D random walks having the same physical time limit with different $\dt$'s.
  }
  \vspace{-15pt}
  \label{fig:dt_exploration}
\end{wrapfigure}
\textbf{Challenging exploration.}
Furthermore, existing PG methods are prone to perform worse with a low $\dt$ due to the difficulty of exploration.
The example in \Cref{fig:dt_exploration} compares 2-D random walks of the same physical time limit with different $\dt$'s.
It is easier to get out of the blue box by chance with a high $\dt$ than a low $\dt$.
Intuitively, when $\dt \rightarrow 0$, the range that the agent can move at each step becomes smaller, which makes it challenging to reach distant states by pure exploration.
This originates again from independently sampled actions.
We provide a more formal explanation in \Cref{sec:ex_exploration}.
This exploration problem has also been addressed in Q-learning literature \citep{ddpg_lillicrap2016,dau_tallec2019},
which employs as a remedy autocorrelated noise such as an Ornstein-Uhlenbeck process.

\textbf{Infinite decision steps.}
For a given time limit, the number of decision steps is inversely proportional to $\dt$.
This makes the size of the training data required for consuming the same number of episodes increase infinitely as $\dt \rightarrow 0$,
and thereby impedes training in terms of computational cost.

\subsection{Durative Actions in Previous Works}
\label{sec:durative}

For the aforementioned problems with PG methods given a low $\dt$, \textit{durative actions} can be a solution;
that is, the policy decides actions only when it is necessary, rather than at every $\dt$.
With durative actions, (1) it naturally makes ($\dt$-discretized) actions correlated with one another,
and (2) it does not always require infinitely many decision steps even if $\dt \to 0$.

For durative actions, one may modify existing policies to produce both action $a$ and its duration $\du$.
In discretized continuous-time environments, this approach is practically equivalent to prior methods that produce repetition counts \citep{figar_sharma2017,dfdqn_lakshminarayanan2017},
as continuous durations should be converted into the repetitions of time-discretized actions.
As one of such action repetition methods, FiGAR-A3C \citep{figar_sharma2017} defines the policy as $\pi(a, x|s)$ where $a \in \mathcal{A}$ and $x \in W$.
$W$ is a predefined set of action repetition counts; \eg, $W = \{1, 2, \ldots, 30\}$.
At every decision step, FiGAR-A3C samples action $a_i$ and repetition count $x_i$ from the policy,
and then performs the action $x_i$ times, where $i$ denotes the $i$-th \textit{decision} step. %
It defines the $n$-step return $\hat{V}^{(n)}(s_i)$ as 
\begin{align}
    \hat{V}^{(n)}(s_i) = \sum_{k=0}^{n-1} \gamma^{y_{i+k} - y_i} r_{i+k} + \gamma^{y_{i+n} - y_i} V(s_{i+n}) \label{eq:discount},
\end{align}
where the cumulative repetition counts $\{y_i\}$ is defined as $y_0 = 0$ and $y_{i+1} = y_i + x_i$ for $i \geq 0$.
The reward $r_i$ at the $i$-th \textit{decision} step is given by the discounted sum of environment rewards over the holding time.
While FiGAR-A3C is based on the standard procedure of A3C \cite{a2c_mnih2016}, it is also applicable to other RL algorithms such as TRPO \citep{trpo_schulman2015} and DDPG \citep{ddpg_lillicrap2016}.

FiGAR can be naturally extended to its continuous variant, which we call FiGAR-C,
by replacing $\pi(a, x|s)$ with $\pi(a, \du|s)$ where $\du \in [0, \du_{\tmax}]$ stands for the \textit{duration} of the action $a$.
In $\dt$-discretized environments,
it translates the action duration $\du$ into $\left\lceil \frac{\du}{\dt} \right\rceil$ repetition times.
FiGAR-C is inherently $\dt$-invariant since it operates on the unit of physical time instead of the discretized time scale.

However, these approaches to action duration have two limitations.
First, since it does not consider stopping an action during repetition,
it cannot immediately react to unexpected events while repeating an action.
This may lead to poor performance in stochastic environments.
Second, in contrast to the fact that the optimal policy $\pi(a|s)$ of a fully observable MDP only depends on states $s$, not time $t$ \citep{rl_sutton2018},
previous methods \textit{only} consider the locality of the time variable $t$, without caring about underlying changes in $s$.
This discrepancy can lead to performance degradation,
since even during a small time period, $s$ can greatly vary in environments with stochastic or non-continuous dynamics.
In the next section, we will demonstrate that
such limitations can still bring back the variance explosion problem in these action repetition approaches.

\subsection{Safe Action Repetition}
\label{sec:sar}

We propose an alternative approach named \textit{Safe Action Repetition} (SAR) that resolves the limitations of existing action repetition methods.
SAR repeats actions based on \textit{state} locality, taking the same action only when states are close.

Following \cite{smooth_shen2020}, we define a \textit{perturbation set} for a state $s \in \mathcal{S}$ as
$\mathbb{B}_{\Delta}(s, \sr) = \{s' | \Delta(s, s') \leq \sr\}$,
which corresponds to the closed ball of radius $\sr$ centered at $s$ using the metric $\Delta$ in the state space.
Employing the notion of perturbation sets, we propose the following action repetition scheme.
At every decision step, SAR's policy $\pi(a_i, \sr_i|s_i)$ produces
both action $a_i$ and the radius $\sr_i$ of a perturbation set,
which we call the \textit{safe region}.
Then, SAR repeats the action $a_i$ only within the safe region $\mathbb{B}_{\Delta}(s_i, \sr_i)$, or equivalently
\begin{align}
    \Delta(s, s_i) \leq \sr_i, \label{eq:sar}
\end{align}
where $s$ denotes the current state during repetition.
Once the agent goes outside of the safe region, SAR stops action repetition and selects a new action.
Having action durations thresholded by \Cref{eq:sar} grants two advantages.
First, it naturally ensures $\dt$-invariance since action duration is determined by the safe region radius,
which is not related to how fine the discretization time scale is.
\Cref{eq:sar} is even completely agnostic to the physical time variable $t$.
Second, the agent becomes robust to stochasticity, \eg, encountering an unpredicted event,
because such a situation would push the agent's state far away from the safe region, which immediately stops action repetition.

\begin{wrapfigure}[10]{r}{0.52\textwidth}
  \centering
  \hspace*{-6.5pt}
  \raisebox{0pt}[\dimexpr\height-0.8\baselineskip\relax]{
    \includegraphics[width=\linewidth]{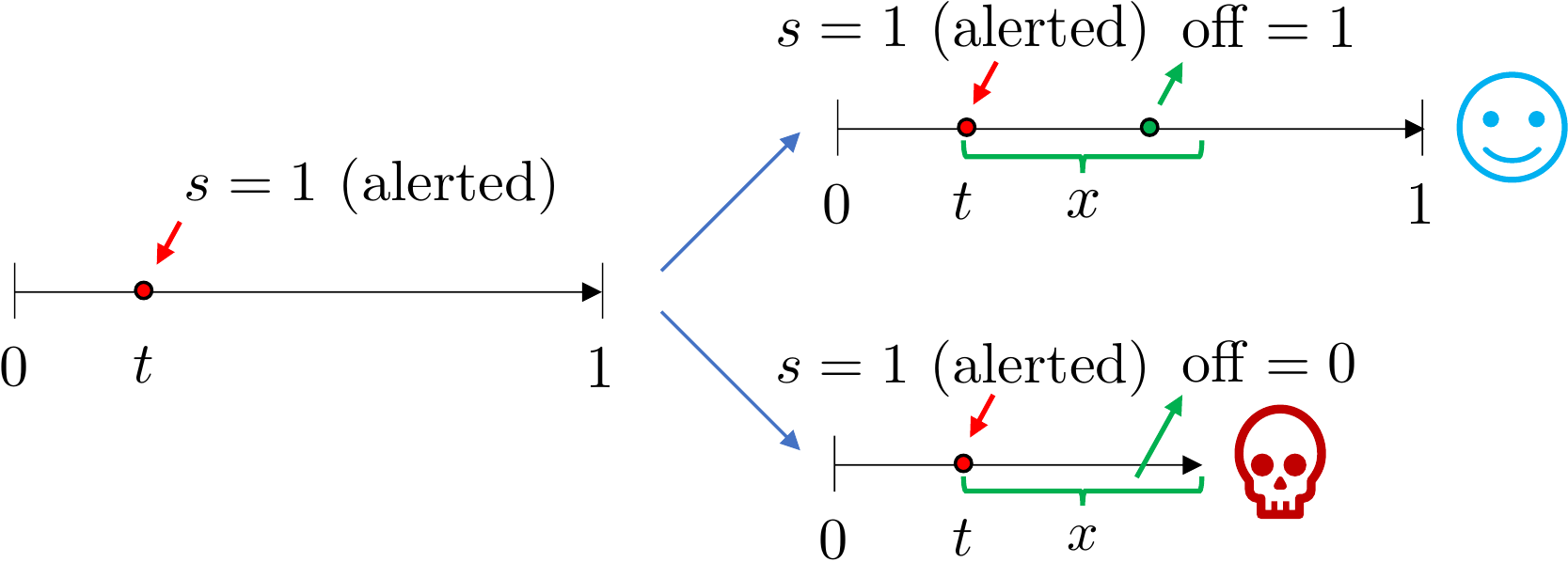}
  }
  \caption{
      Illustration of \textit{AlertThenOff} environment.
  }
  \label{fig:ex_env}
\end{wrapfigure}
To intuitively differentiate SAR with previous action repetition methods,
we illustrate a simple example of a stochastic environment, where previous FiGAR-C fails to maintain an optimal policy due to the variance explosion of the PG estimator, while our method does not.
Let us consider the following $\dt$-discretized environment named \textit{AlertThenOff}, whose physical time limit $T$ is $1$.
The state is 
$s \in \{0\ (\text{normal}),\ 1\ (\text{alerted})\}$ with $s_0 = 0\ (\text{normal})$.
The 2-D action is $a = [\mathrm{off}, \mathrm{num}]^\top$
where %
$\mathrm{off} \in \{0, 1\}$ and $\mathrm{num} \in \mathbb{R}$.
In this environment, $s$ randomly changes once from $0\ (\text{normal})$ to $1\ (\text{alerted})$ at time $t \in [0, 1]$.
When $s=1\ (\text{alerted})$, we should take an action of $\mathrm{off} = 1$
within $[t, t+x]$ so that $s$ can be back to $0\ (\text{normal})$,
where $x > \dt > 0$ is an environment parameter known to us.
Otherwise, the environment produces a penalty reward of $-\nu$ (a large negative number) and ends immediately.
Conversely, if we set $\mathrm{off} = 1$ when $s = 0\ (\text{normal})$, it also causes a penalty of $-\nu$ and ends immediately.
We illustrate this in \Cref{fig:ex_env}.
The reward (before discretization) is given by $r(t) = f(\mathrm{num})$, where $f$ is an unknown reward function,
and its discretized reward is given accordingly to \Cref{eq:delta_reward}.
When time reaches $t=1$, a noisy reward of $\xi \sim \mathcal{N}(0, 1)$ occurs.
To sum up, we should perform $\mathrm{off} = 1$ as soon as possible
only when we get to know that $s$ becomes $1\ (\text{alerted})$,
while we should also perform appropriate $\mathrm{num}$ actions that maximize $f(\mathrm{num})$.

To find an optimal policy in this environment,
we consider a deterministic policy for $\mathrm{off}$ such that $\pi^{\mathrm{off}}(s) = s$,
which we already know is optimal,
and a stochastic policy for $\mathrm{num}$ that $\pi^{\mathrm{num}}(\mathrm{num}|s) \sim \mathcal{N}(\mu, 1)$,
where $\mu$ is the policy's parameter.
Our goal is thus to find the optimal value of $\mu$.
If we assume that the penalty is infinitely large and $f \equiv 0$ for simplicity,
we obtain the following result.

\begin{proposition}
    \label[proposition]{th:example}
    In AlertThenOff environment,
    for the optimal policy $\pi_{\theta_f}$ for FiGAR-C
    and the optimal policy $\pi_{\theta_s}$ for SAR,
    the following holds:
    \begin{align}
        \mathrm{tr} \left[ \mathbb{V}_{\tau \sim p_{\theta_f}(\tau)}[G_{\theta_f}(\tau)] \right] \to \infty, \hspace{12pt}
        \mathrm{tr} \left[ \mathbb{V}_{\tau \sim p_{\theta_s}(\tau)}[G_{\theta_s}(\tau)] \right] = 2,
    \end{align}
    when $\dt \to 0$, $x \to 0$, $\nu \to \infty$ and $f \equiv 0$.
\end{proposition}

We provide a proof in \Cref{sec:proof2}.
From \Cref{th:example}, we can conclude that in contrast to FiGAR-C,
our SAR policy does not suffer from variance explosion in \textit{AlertThenOff} environment.
As the previous approach fails to maintain an optimal policy even in this very simple stochastic environment,
it can also be at risk of failure in general stochastic environments.
The intuition behind this failure is that it has no choice but
to infinitely shorten action durations in order to optimally handle stochasticity that requires immediate reactions from the agent,
which leads both the number of decision steps and the variance of the policy gradient to explode.
On the other hand,
SAR can be free from variance explosion despite such stochasticity,
if SAR sets appropriate safe regions so that such an exigent state locates outside of the safe regions,
and thereby the number of decision steps can be bounded.

\subsection{SAR on Non-Markovian Environments}
\label{sec:sar_nme}

We derived SAR based on the fact that the optimal policy in a fully observable MDP only depends on states.
However, if the Markovian property does not hold (\eg, environments with partially observable MDPs \citep{drqn_hausknecht2015} or time limits \citep{time_pardo2018}),
the optimal policy might not be fully determined by states alone.
In this case, we can additionally incorporate temporal thresholds into SAR
by extending the definition of the safe region as follows:
\begin{align}
    \lambda {\cdot} \Delta(s, s_i) + (1 - \lambda) |t - t_i| \leq \sr_i,
\end{align}
where $t_i$ denotes the time at the $i$-th step, $t$ denotes the current time during repetition,
and $0 \leq \lambda \leq 1$ is the coefficient that controls the trade-off between distance and time differences.
Note that this variant of SAR, which we call \textit{$\lambda$-SAR}, is $\dt$-invariant too
since each term is independent from $\dt$.

\section{Experiments}
\label{sec:experiments}

We apply SAR to three policy gradient (PG) methods, PPO \citep{ppo_schulman2017}, TRPO \citep{trpo_schulman2015} and A2C \citep{a2c_mnih2016},
and compare with baseline methods in multiple settings.
We first demonstrate the $\dt$-invariance of our method on deterministic continuous control environments,
compared to previous $\dt$-invariant algorithms
(\Cref{sec:exp_determ}). %
We then evaluate SAR on stochastic environments to show its robustness to stochasticity (\Cref{sec:exp_stoch}).
We also provide illustrative examples for a better understanding of our method (\Cref{sec:exp_illust}).
We describe the full experimental details in \Cref{sec:experimental_details}.

\textbf{Experimental setup.}
We test SAR on eight continuous control environments from MuJoCo \citep{mujoco_todorov2012}:
InvertedPendulum-v2, InvertedDoublePendulum-v2, Hopper-v2, Walker2d-v2,
HalfCheetah-v2, Ant-v2, Reacher-v2 and Swimmer-v2.
We mainly compare our method to FiGAR-C described in \Cref{sec:durative}
because it is the only prior method that is $\dt$-invariant and does not always require infinite decision steps even if $\dt \to 0$,
but we also make additional comparisons with other baselines
such as DAU \citep{dau_tallec2019}, ARP \citep{ar_korenkevych2019}, modified PPO as well in \Cref{sec:exp_determ} and \Cref{sec:arp,sec:further_ppo}.

For SAR's distance function in \Cref{eq:sar}, we use
$\Delta(s, s_i) = \| \tilde{s} - \tilde{s_i} \|_1 / \mathrm{dim}(\mathcal{S})$,
where $\|\cdot\|_1$ is the $\ell_1$ norm and $\tilde{s}$ is the state normalized by its moving average.
This distance function corresponds to the average difference in each normalized state dimension,
where the normalization permits sharing the hyperparameter $\sr_\tmax$ for all MuJoCo tasks.
We also share $\du_\tmax$ in FiGAR-C for all environments.
Finally, we impose an upper limit of $\du_\tmax$ on the maximum duration of actions in SAR for two reasons:
(1) to further stabilize training and
(2) to ensure a fair comparison with FiGAR-C by setting the same limit on time duration.
We provide an ablation study including an analysis of imposing an upper limit on $t$ in \Cref{sec:abl}.

\subsection{Results on Deterministic Environments}
\label{sec:exp_determ}

\begin{figure}[t!]
  \centering
  \includegraphics[width=\linewidth]{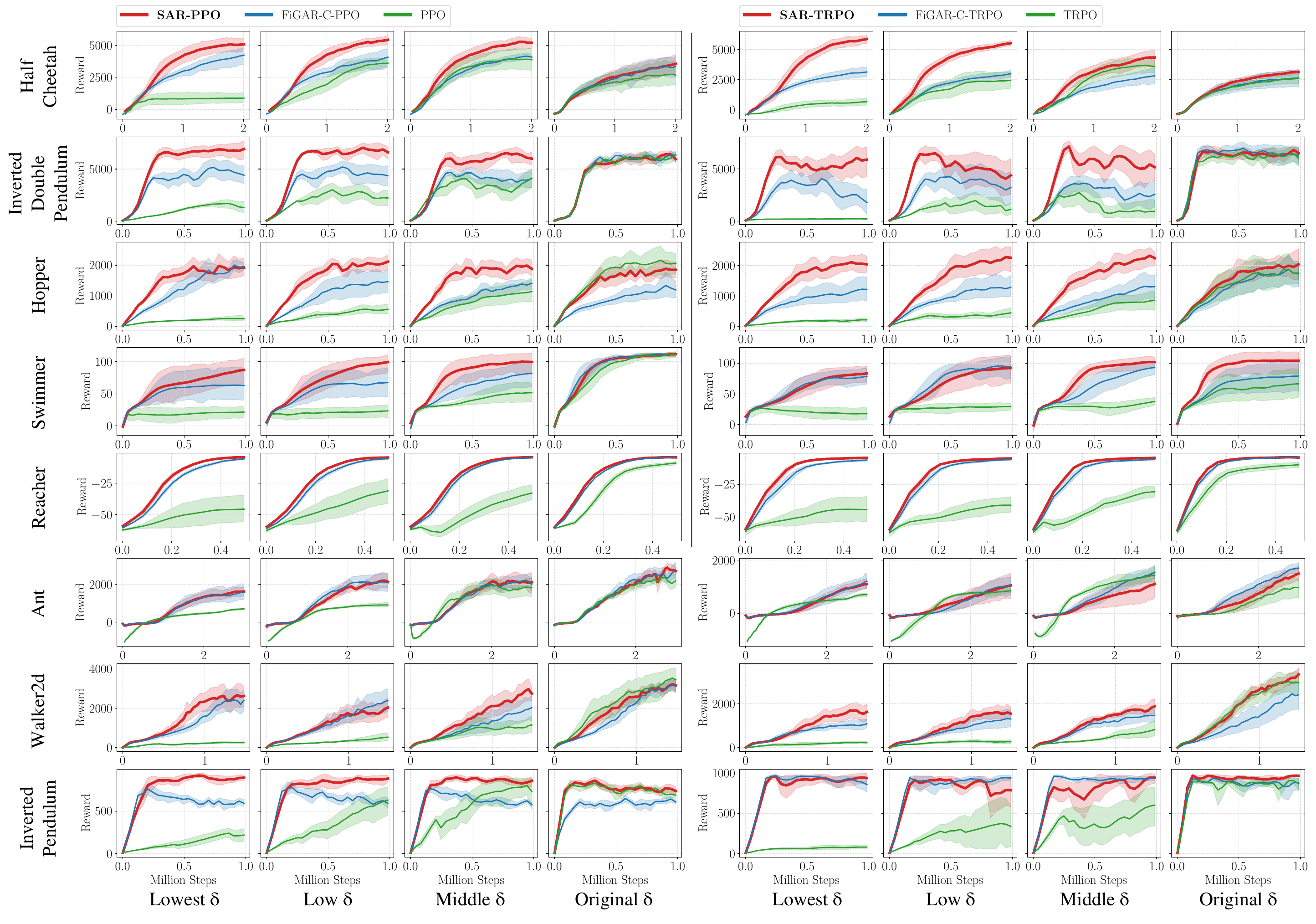}
  \caption{
      Training curves on deterministic MuJoCo environments with various $\dt$'s (ranging from $5e-4$ to $5e-2$).
      Shaded areas represent the $95\%$ confidence intervals over eight runs.
      We compare SAR to FiGAR-C with two base PG algorithms, PPO and TRPO.
      SAR mostly has $\dt$-invariance, showing similar or even better performance with lower $\dt$'s. %
  }
  \label{fig:determ}
\end{figure}

We first train SAR and FiGAR-C with PPO, TRPO and A2C on the MuJoCo environments, which have deterministic transition dynamics (although they have randomized initial states),
with various discretization time scales ranging from $5e-4$ to $5e-2$ (details in \Cref{sec:experimental_details}).
We use suffixes such as `-PPO' to denote the base PG algorithms.
\Cref{fig:determ} shows the training curves of PPO and TRPO with SAR and with FiGAR-C on the eight MuJoCo environments,
where the $x$ and $y$ axes denote the number of decision steps and the total reward, respectively.
We provide results on A2C and further comparison between SAR-PPO and PPO in \Cref{sec:add_results,sec:further_pg}.
\Cref{fig:determ} shows that while vanilla PG algorithms fail to maintain their performance in lower-$\dt$ settings,
SAR is mostly robust to varying $\dt$, often even achieving the best performance with the lowest $\dt$.
In a few environments such as Swimmer-v2, SAR's performance becomes slightly worse as $\dt$ decreases,
which we hypothesize is because a higher $\dt$ aids the agent in getting out of local optima.
Also, SAR exhibits similar or better performance compared to FiGAR-C on most deterministic environments.

\begin{figure}[t!]
  \centering
  \includegraphics[width=\linewidth]{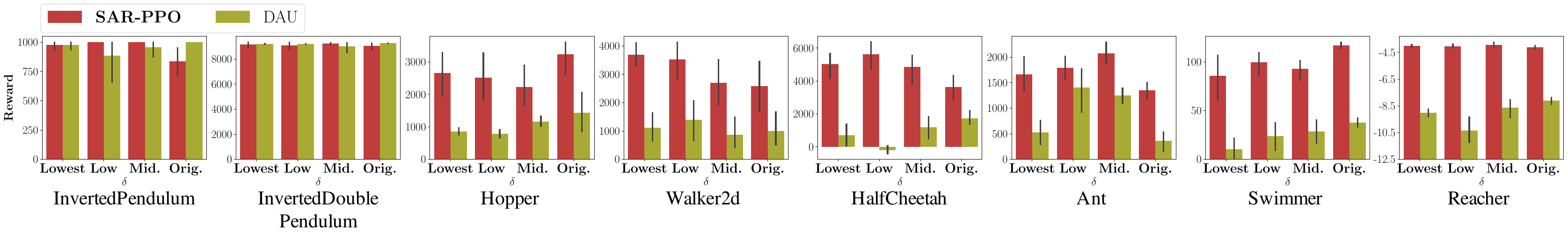}
  \caption{
      Bar plots comparing the final performance of SAR-PPO to DAU's on deterministic MuJoCo environments with various $\dt$'s.
      Error bars represent the $95\%$ confidence intervals over eight runs.
  }
  \label{fig:dau}
\end{figure}

For a more comprehensive evaluation,
we additionally compare with DAU \citep{dau_tallec2019},
which is another approach to $\dt$-invariance for Q-learning methods such as DQN \citep{dqn_mnih2013} and DDPG \citep{ddpg_lillicrap2016}.
Note that SAR-PPO and DAU have different underlying algorithms and training schemes.
While DAU is based on DDPG and chooses an action at every environment step,
SAR-PPO operates on PPO and makes action decisions only when needed.
For the comparison, we use the official implementation of DAU \citep{dau_tallec2019}.
\Cref{fig:dau} compares the final performances of both methods with various $\dt$'s
at the same physical time (equal to $1e6$ environment steps in the original $\dt$) on each environment.
Overall SAR-PPO outperforms DAU and exhibits better $\dt$-invariance in most of the environments.
Also, our method requires about $17.6\times$ fewer decision steps on average in the lowest-$\dt$ settings via temporally extended actions.

\subsection{Results on Stochastic Environments}
\label{sec:exp_stoch}

\begin{figure}[t!]
  \centering
  \includegraphics[width=\linewidth]{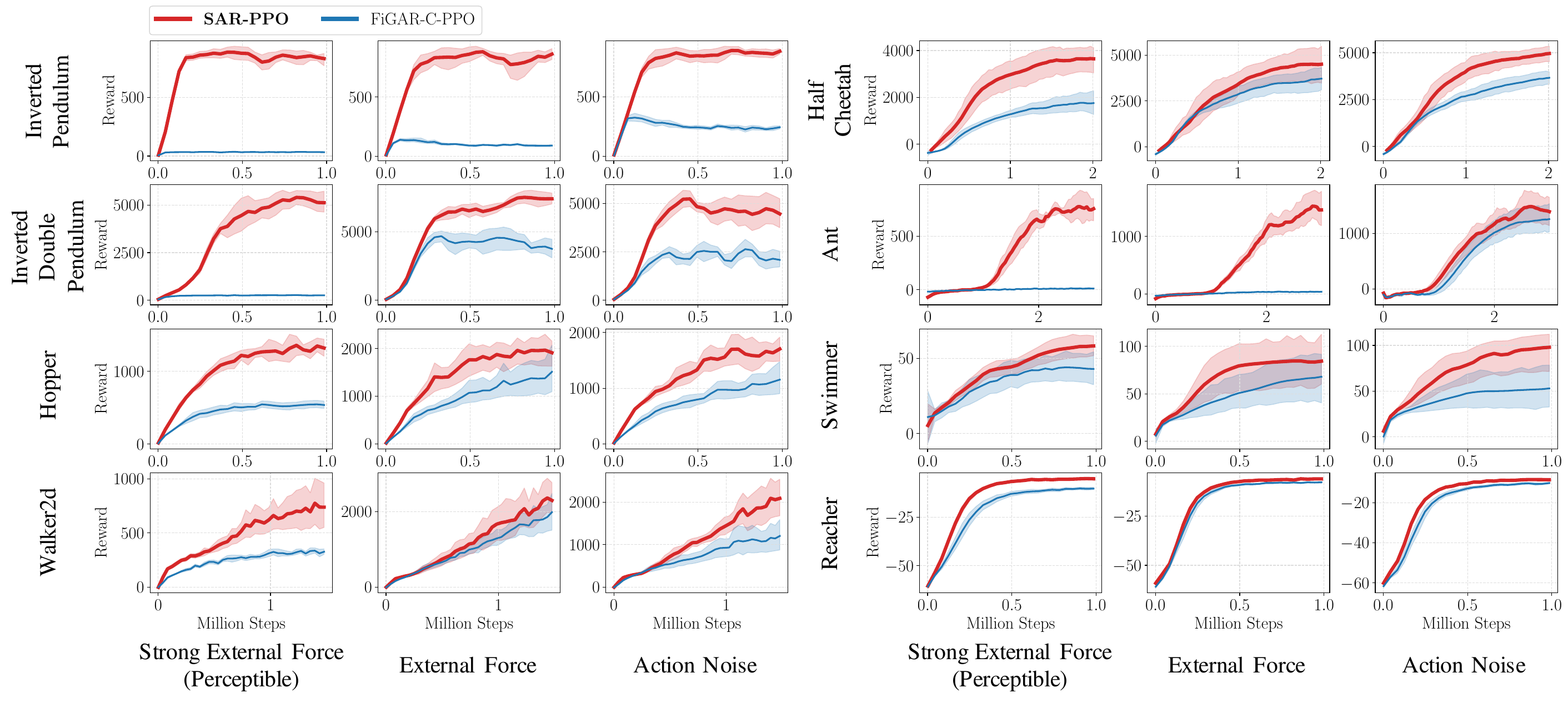}
  \caption{
      Training curves of SAR-PPO and FiGAR-C-PPO on MuJoCo environments with various types of stochasticity.
      Shaded areas represent the $95\%$ confidence intervals over eight runs.
      SAR exhibits strong performance in the presence of stochasticity.
  }
  \label{fig:stoch_ppo}
\end{figure}

To demonstrate SAR's robustness with stochastic dynamics,
we modify existing MuJoCo environments by adding various types of stochasticity.
(1) ``External Force'': we apply an external force with a standard deviation of $\sigma_{\text{ext}}$ to the agent's body
with a probability of $p_{\text{ext}}$ at each decision step.
(2) ``Strong External Force (Perceptible)'':
we make external forces perceptible by the agent,
which allows it to react to stronger forces with $\sigma_{\text{ext2}} > \sigma_{\text{ext}}$.
(3) ``Action Noise'': we apply noise with a standard deviation of $\sigma_{\text{act}}$ to the action
with a probability of $p_{\text{act}}$ at each decision step.
Throughout this experiment, we use the lowest-$\dt$ settings.
\Cref{fig:stoch_ppo} compares SAR-PPO's performance with FiGAR-C-PPO's, and shows that SAR outperforms FiGAR-C on most of the stochastic environments, often exhibiting drastic differences.
We provide further details and the comparison on TRPO in \Cref{sec:add_results,sec:experimental_details}.

\subsection{Qualitative Analysis of SAR}
\label{sec:exp_illust}

\begin{figure}[t!]
  \centering
  \begin{subfigure}[t]{0.97\linewidth}
    \includegraphics[width=\linewidth]{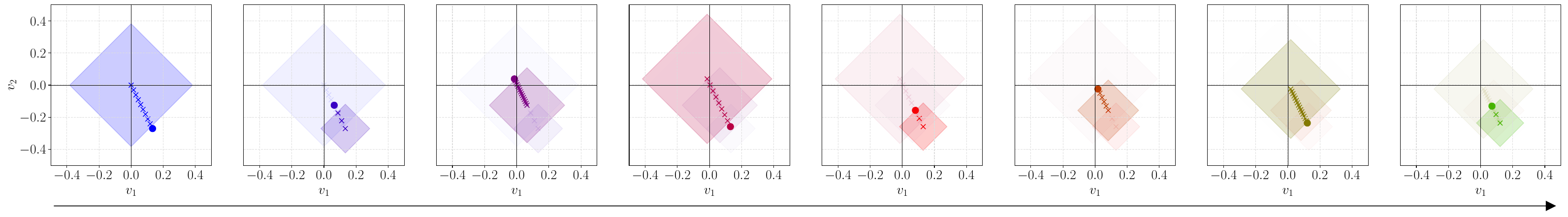}
    \caption{InvertedPendulum-v2}
    \label{fig:inp_analysis}
  \end{subfigure}
  \begin{subfigure}[t]{0.97\linewidth}
    \includegraphics[width=\linewidth]{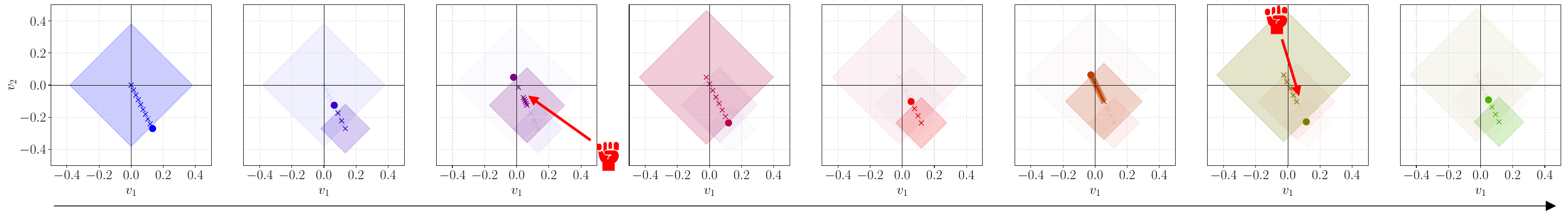}
    \caption{InvertedPendulum-v2 with external forces}
    \label{fig:inp_analysis_noi}
  \end{subfigure}
  \caption{
      Illustration of SAR on InvertedPendulum-v2 with $\dt = 2e-3$.
      Our policy produces safe regions (rhombus) within which the agent repeats actions.
      Circle markers represent the \textit{break points} of action repetitions and red fists represent the external forces applied to the agent.
  }
\end{figure}

In order to provide further insights on SAR, we illustrate how SAR works on InvertedPendulum-v2,
where the goal is to maintain the balance of a pendulum. %
The state space consists of four dimensions: two for the position of the agent and the other two for its velocity;
\ie, $s = [x_1, x_2, v_1, v_2]$.
We demonstrate the behavior of SAR on a 2-D plane.
For better interpretability, we slightly modify our method in this experiment:
we use only the two (normalized) velocity dimensions for the distance function;
that is, the agent stops action repetitions only by its velocity.

\Cref{fig:inp_analysis} illustrates how a trained SAR model performs action repetition based on safe regions,
where the $x$ and $y$ axes correspond respectively to $v_1$ and $v_2$,
rhombuses represent safe regions,
markers (either circle or cross) represent time-discretized states.
It can be observed that our policy is learned to produce a large safe region when the agent's speed is low
and, conversely, a small safe region when the speed is high,
which fits with the intuition because the risk of losing the balance of the pendulum rises as the agent's speed increases.

Additionally, we demonstrate how SAR operates in the presence of stochastic external forces (described in \Cref{sec:exp_stoch}) in \Cref{fig:inp_analysis_noi}.
It shows how SAR can quickly and adaptively handle such stochasticity with safe regions by immediately stopping repetitions.

\textbf{How SAR learns the adaptive sizes of safe regions?}
When the agent's velocity is low,
a bigger safe region becomes a low-hanging fruit especially in the early stages of the training
where the PG estimator is mostly dominated by immediate rewards
(\ie, the accumulated reward within a single action repetition),
and thus SAR gravitates toward producing larger safe regions.
On the other hand, when the velocity is high and if safe regions are too large,
the pendulum would easily lose the balance and thus lead to the end of the episode,
which makes SAR in favor of smaller safe regions.

\section{Conclusion}
\label{sec:conclusion}
We proposed Safe Action Repetition (SAR), a novel $\dt$-invariant action repetition method for policy gradient (PG) algorithms.
SAR can handle infinitesimal-$\dt$ settings using temporally extended actions without suffering from the variance explosion problem of PG methods,
which we proved for general stochastic environments under a certain assumption.
It can agilely cope with environment stochasticity via learned safe regions.
We exhibited that our method achieves both $\dt$-invariance and robustness to stochasticity.

\textbf{Limitations and future directions.}
SAR operates in environments where the distance functions can be properly defined in the state spaces.
We experimented with the $\ell_1$ norm as SAR's distance function,
which might not be applicable to discrete or very high dimensional environments such as vision-based simulations.
As such, we expect that combining our method with representation learning techniques
\citep{dreamer_hafner2020,dreamerv2_hafner2021,muzero_schrittwieser2020} would be an interesting future research direction.
Also, since SAR in \Cref{sec:sar} assumes fully observable MDPs so that it can exploit state locality,
there is room for improvement in environments with noisy states or partially observable MDPs.
Although we have suggested one possible solution in \Cref{sec:sar_nme}, this could also be the subject of future research.

\section{Broader Impact}
\label{sec:broader}

We expect that our method is especially useful
in a variety of real-world situations
where the sampling frequencies of simulated environments and real-world physical sensors are different
or where the agent could encounter unexpected situations in the environment.
However, in spite of the potential positive aspects,
practitioners need to pay sufficient attention to various perspectives on their problems and our assumptions
when trying to apply our proposed method to real-world problems.
For example, in some environments such as autonomous driving, minimizing risk may be more crucial than maximizing rewards, for which 
they may need to consider incorporating risk-averse methods \cite{risk_rockafellar2002,risk_singh2020}.
Also, they have to examine the degree to which the assumptions made in this work are satisfied in their problems.
For instance, the Markovian property may not generally hold in practical settings, in which applying our method as-is might possess potential risks.
They should analyze the ramifications of the assumption mismatch and handle them accordingly (see also \Cref{sec:conclusion}).
With such considerations, we hope that our research provides new insights toward $\dt$-invariant and robust reinforcement learning.

\section*{Acknowledgements}

We thank the anonymous reviewers for their helpful comments.
This work was supported by 
Samsung Advanced Institute of Technology,     %
Brain Research Program by National Research Foundation of Korea (NRF) (2017M3C7A1047860),     %
Institute of Information \& communications Technology Planning \& Evaluation (IITP) grant funded by the Korea government (MSIT) (No.2019-0-01082, SW StarLab) and     %
Institute of Information \& communications Technology Planning \& Evaluation (IITP) grant funded by the Korea government (MSIT) (No.2021-0-01343, Artificial Intelligence Graduate School Program (Seoul National University)).
Gunhee Kim is the corresponding author.

\bibliographystyle{plainnat}
\bibliography{action_repetition}

\appendix

\clearpage

\section{Proof of Theorem 1}
\label{sec:proof1}

In this section, we derive a lower bound for the trace of the covariance of the PG estimator in environments with stochastic dynamics.
Recall that 
$\pi_\theta(a_i|s_i) \sim \mathcal{N}(\mu_{\theta_\mu}(s_i), \Sigma)$
and $p_\theta(\tau) = p(s_0)\prod_{i=0}^{N-1} \pi_\theta(a_i|s_i)p(s_{i+1}|s_i,a_i)$
,
where the mean $\mu_{\theta_\mu}(s_i) = [\mu_{\theta_\mu,1}(s_i), \ldots, \mu_{\theta_\mu,K}(s_i)]^\top$
(the subscript $k$ in $\mu_{\theta_\mu,k}(s_i)$ denotes the $k$-th element of the vector $\mu_{\theta_\mu}(s_i)$)
is modeled by a neural network,
$\Sigma = \mathrm{diag}(\sigma^2_1, \ldots, \sigma^2_K)$ is the learnable variance that is independent of states,
$\theta = [\theta_\mu^\top, \sigma_1, \ldots, \sigma_K]^\top$ is the whole parameters of the policy $\pi_\theta$,
and $K = \mathrm{dim}(\mathcal{A})$.

First, let us consider a single scalar parameter $\vartheta$ that is a bias in the last layer of the mean network $\mu_{\theta_\mu}$,
such that $\mu_{\theta_\mu}(s_i) = \mu'_{\theta_{\mu'}}(s_i) + [b_1, b_2, \ldots, b_K]^\top$ and
w.l.o.g.
$\sigma_1 = \mathrm{min}(\sigma_1, \sigma_2, \ldots, \sigma_K)$
and $\vartheta \triangleq b_1$,
where $[b_1, \ldots, b_K]$ is the set of bias parameters in the last layer and
$\mu'_{\theta_{\mu'}}(s_i)$ denotes the remainder of the mean network.
Then, the following holds:
\begin{align}
    &\mkern-18mu \mathrm{tr} \left[ \mathbb{V}_{\tau \sim p_\theta(\tau)}
    \left[ \left( \sum_{i=0}^{N-1} \nabla_\theta \log \pi_\theta (a_i | s_i) \right) R(\tau) \right] \right]\\
    &\geq \mathbb{V}_{\tau \sim p_\theta(\tau)}
    \left[ \left( \sum_{i=0}^{N-1} \frac{\partial}{\partial\vartheta} \log \pi_\theta (a_i | s_i) \right) R(\tau) \right]\\
    &= \mathbb{V}_{\tau \sim p_\theta(\tau)}
    \left[ \left( \sum_{i=0}^{N-1}
    \frac{\partial}{\partial\vartheta} \sum_{k=1}^K
    \left(-\frac{1}{2\sigma^2_k} (a_{i,k} - \mu_{\theta_\mu,k}(s_i))^2 - \frac{1}{2}\log(2\pi\sigma^2_k)\right)
    \right) R(\tau) \right]\\
    &= \mathbb{V}_{\tau \sim p_\theta(\tau)}
    \left[ \left( \sum_{i=0}^{N-1}
    \sum_{k=1}^K \frac{1}{\sigma^2_k} (a_{i,k} - \mu_{\theta_\mu,k}(s_i)) \frac{\partial}{\partial\vartheta} \mu_{\theta_\mu,k}(s_i)
    \right) R(\tau) \right]\\
    &= \mathbb{V}_{\tau \sim p_\theta(\tau)}
    \left[ \left( \sum_{i=0}^{N-1}
    \frac{1}{\sigma^2_1} (a_{i,1} - \mu_{\theta_\mu,1}(s_i))
    \right) R(\tau) \right].
\end{align}

By reparameterizing the actions as $a_{i,k} = \mu_{\theta_\mu,k}(s_i) + \sigma_{k}\epsilon_{i,k}$
for all $0 \leq i \leq N-1$ and $1 \leq k \leq K$,
where $\{\epsilon_{i,k}\} \overset{\text{i.i.d.}}{\sim} \mathcal{N}(0, 1)$
(we use the simplified notation $\{\epsilon_{i,k}\}$ to denote $(\epsilon_{0,1}, \ldots, \epsilon_{0,K}, \epsilon_{1,1}, \ldots, \epsilon_{N-1,K})$), we obtain
\begin{align}
    &\mkern-18mu \mathbb{V}_{\tau \sim p_\theta(\tau)}
    \left[ \left( \sum_{i=0}^{N-1} \frac{1}{\sigma^2_1} (a_{i,1} - \mu_{\theta_\mu,1}(s_i)) \right) R(\tau) \right]\\
    &=\mathop{\mathbb{V}}_{\substack{\{\epsilon_{i,k}\} \overset{\text{i.i.d.}}{\sim} \mathcal{N}(0, 1)
    \\ s_{0:N} \sim p_\theta(s_{0:N}|\{\epsilon_{i,k}\})}}
    \left[ \left( \sum_{i=0}^{N-1} \frac{\epsilon_{i,1}}{\sigma_1} \right) R(\tau) \right] \label{eq:before_tv}.
\end{align}

We then use the law of total variance to decompose \Cref{eq:before_tv} as follows:
\begin{align}
    &\mkern-18mu \mathop{\mathbb{V}}_{\substack{\{\epsilon_{i,k}\} \overset{\text{i.i.d.}}{\sim} \mathcal{N}(0, 1)
    \\ s_{0:N} \sim p_\theta(s_{0:N}|\{\epsilon_{i,k}\})}}
    \left[ \left( \sum_{i=0}^{N-1} \frac{\epsilon_{i,1}}{\sigma_1} \right) R(\tau) \right]\\
    &=\mathbb{V}_{\{\epsilon_{i,k}\} \overset{\text{i.i.d.}}{\sim} \mathcal{N}(0, 1)}\left[
        \mathbb{E}_{s_{0:N} \sim p_\theta(s_{0:N}|\{\epsilon_{i,k}\})}\left[
            \left( \sum_{i=0}^{N-1} \frac{\epsilon_{i,1}}{\sigma_1} \right) R(\tau)
        \right]
    \right]\nonumber\\
    &\quad+\mathbb{E}_{\{\epsilon_{i,k}\} \overset{\text{i.i.d.}}{\sim} \mathcal{N}(0, 1)}\left[
        \mathbb{V}_{s_{0:N} \sim p_\theta(s_{0:N}|\{\epsilon_{i,k}\})}\left[
            \left( \sum_{i=0}^{N-1} \frac{\epsilon_{i,1}}{\sigma_1} \right) R(\tau)
        \right]
    \right]\\
    &\geq \mathbb{E}_{\{\epsilon_{i,k}\} \overset{\text{i.i.d.}}{\sim} \mathcal{N}(0, 1)}\left[
        \mathbb{V}_{s_{0:N} \sim p_\theta(s_{0:N}|\{\epsilon_{i,k}\})}\left[
            \left( \sum_{i=0}^{N-1} \frac{\epsilon_{i,1}}{\sigma_1} \right) R(\tau)
        \right]
    \right]\\
    &=\mathbb{E}_{\{\epsilon_{i,k}\} \overset{\text{i.i.d.}}{\sim} \mathcal{N}(0, 1)}\left[
        \frac{(\sum_{i=0}^{N-1} \epsilon_{i,1})^2}{\sigma^2_1}
        \mathbb{V}_{s_{0:N} \sim p_\theta(s_{0:N}|\{\epsilon_{i,k}\})}\left[ R(\tau) \right]
    \right] \label{eq:before_assume}.
\end{align}

Now we apply the following assumption:
\begin{align}
    \forall \{\epsilon_{i, k}\} \quad
    \mathbb{V}_{s_{0:N} \sim p_\theta(s_{0:N}|\{\epsilon_{i,k}\})}\left[ R(\tau) \right] \geq c,
\end{align}
where $c$ is a small constant greater than 0.
This assumption states that the environment is inherently stochastic in the sense that
its return has a variance of at least $c$ even if conditioned on the reparameterized actions $\{\epsilon_{i, k}\}$
(or equivalently, $\mathbb{V}[R(\tau)]$ is greater than or equal to $c$ even if the \emph{random seed} used for sampling actions is fixed).
Note that environments with deterministic transition dynamics, such as the MuJoCo environments,
could also satisfy this assumption, considering the stochasticity of the initial state: $s_0 \sim p(s_0)$
(although a perfect baseline could cancel out the initial stochasticity in such environments).

Using this assumption, \Cref{eq:before_assume} can be rewritten as
\begin{align}
    &\mkern-18mu \mathbb{E}_{\{\epsilon_{i,k}\} \overset{\text{i.i.d.}}{\sim} \mathcal{N}(0, 1)}\left[
        \frac{(\sum_{i=0}^{N-1} \epsilon_{i,1})^2}{\sigma^2_1}
        \mathbb{V}_{s_{0:N} \sim p_\theta(s_{0:N}|\{\epsilon_{i,k}\})}\left[ R(\tau) \right]
    \right]\\
    &\geq \mathbb{E}_{\{\epsilon_{i,k}\} \overset{\text{i.i.d.}}{\sim} \mathcal{N}(0, 1)}\left[
        \frac{(\sum_{i=0}^{N-1} \epsilon_{i,1})^2}{\sigma^2_1} c
    \right] \label{eq:cause_explosion1}\\
    &=\frac{Nc}{\sigma^2_1} \label{eq:cause_explosion2}\\
    &=\frac{Tc}{\delta\sigma^2_1}\\
    &=\frac{Tc}{\delta \cdot \mathrm{min}(\sigma_1^2, \sigma_2^2, \ldots, \sigma_K^2)} \label{eq:explosion}.
\end{align}

From \Cref{eq:explosion}, we can conclude that $\dt \to 0$ leads the variance of the PG estimator to explode in stochastic environments.

As a side note, if we leave the other term when decomposing \Cref{eq:before_tv}, we obtain
\begin{align}
    &\mkern-18mu \mathop{\mathbb{V}}_{\substack{\{\epsilon_{i,k}\} \overset{\text{i.i.d.}}{\sim} \mathcal{N}(0, 1)
    \\ s_{0:N} \sim p_\theta(s_{0:N}|\{\epsilon_{i,k}\})}}
    \left[ \left( \sum_{i=0}^{N-1} \frac{\epsilon_{i,1}}{\sigma_1} \right) R(\tau) \right]\\
    &=\mathbb{V}_{\{\epsilon_{i,k}\} \overset{\text{i.i.d.}}{\sim} \mathcal{N}(0, 1)}\left[
        \mathbb{E}_{s_{0:N} \sim p_\theta(s_{0:N}|\{\epsilon_{i,k}\})}\left[
            \left( \sum_{i=0}^{N-1} \frac{\epsilon_{i,1}}{\sigma_1} \right) R(\tau)
        \right]
    \right]\nonumber\\
    &\quad+\mathbb{E}_{\{\epsilon_{i,k}\} \overset{\text{i.i.d.}}{\sim} \mathcal{N}(0, 1)}\left[
        \mathbb{V}_{s_{0:N} \sim p_\theta(s_{0:N}|\{\epsilon_{i,k}\})}\left[
            \left( \sum_{i=0}^{N-1} \frac{\epsilon_{i,1}}{\sigma_1} \right) R(\tau)
        \right]
    \right]\\
    &\geq \mathbb{V}_{\{\epsilon_{i,k}\} \overset{\text{i.i.d.}}{\sim} \mathcal{N}(0, 1)}\left[
        \mathbb{E}_{s_{0:N} \sim p_\theta(s_{0:N}|\{\epsilon_{i,k}\})}\left[
            \left( \sum_{i=0}^{N-1} \frac{\epsilon_{i,1}}{\sigma_1} \right) R(\tau)
        \right]
    \right]\\
    &=\mathbb{V}_{\{\epsilon_{i,k}\} \overset{\text{i.i.d.}}{\sim} \mathcal{N}(0, 1)}\left[
        \left( \sum_{i=0}^{N-1} \frac{\epsilon_{i,1}}{\sigma_1} \right)
        \mathbb{E}_{s_{0:N} \sim p_\theta(s_{0:N}|\{\epsilon_{i,k}\})}\left[ R(\tau) \right]
    \right].
\end{align}
From this, we can speculate that even in completely deterministic environments, $\dt \to 0$ is also likely to cause variance explosion
(\eg, consider a simple setting with $R(\tau) = 1$),
but generalizing this may require more sophisticated assumptions, which we leave for future work.

\section{Concrete Example of the Challenging Exploration Problem}
\label{sec:ex_exploration}

In this section, we illustrate the difficulty of exploration with a low $\dt$.
Let us consider the following simple continuous-time MDP defined as
\begin{align}
    T &= 2\\
    \gamma &= 1\\
    s(t) &\in \mathbb{R}^2\\
    a(t) &\in \{-1, +1\}\\
    s(0) &= [0, 0]^\top\\
    F(s(t), a(t)) &= [a(t), 1]^\top,
\end{align}
where $T$ denotes the physical time limit.

Its discretized MDP with a discretization time scale $\dt = \frac{2}{N}$, which equally divides the total duration by $N$,
is defined as follows:
\begin{align}
    \tau &= (s_0, a_0, \ldots, s_{N})\\
    s_0 &= [0, 0]^\top\\
    s_{i+1} &= s_i + \left[\frac{2}{N}a_i, \frac{2}{N}\right]^\top\\
    r(s_i, a_i) &= \mathds{1}_{\{|s_{i,0}| \geq 1 \: \text{and} \: s_{i,1} \geq 2 \}},
\end{align}
where $\mathds{1}$ denotes the indicator function and we additionally define the reward function $r(s_i, a_i)$.

Let us assume that the initial policy $\pi(a_i|s_i)$ follows the uniform distribution such that $\pi(a_i = -1|s_i) = \pi(a_i = +1|s_i) = \frac{1}{2}$ for all $i$.
Intuitively, this corresponds to a simple 1-D random walk process,
where the first dimension of the state denotes the agent's position, the second dimension denotes the current time, and
a positive reward occurs if the final position $s_{N,0}$ of the agent is located outside of the interval $(-1, 1)$.

When $\dt \to 0$, we can compute the probability that the agent gets a positive reward with the policy $\pi$ as follows:
\begin{align}
    P(R_\pi(\tau) > 0) &= 1 - P\left(\frac{1}{4}N < X < \frac{3}{4}N\right) &&\text{where} \enspace X \sim B\left(N, \frac{1}{2}\right)\\
                       &\approx 1 - P\left(\frac{1}{4}N < Y < \frac{3}{4}N\right) &&\text{where} \enspace Y \sim \mathcal{N}\left(\frac{N}{2}, \frac{N}{4}\right)\\
                       &= 1 - P\left(-\frac{\sqrt{N}}{2} < Z < \frac{\sqrt{N}}{2}\right) &&\text{where} \enspace Z \sim \mathcal{N}(0, 1)\\
                       &\to 0 &&\text{as} \enspace N = \frac{2}{\dt} \to \infty \label{eq:exploration},
\end{align}
where $R_\pi(\tau)$ denotes the random variable corresponding to the return of the trajectory obtained by the policy $\pi$, $B$ denotes the binomial distribution and $\mathcal{N}$ denotes the normal distribution.
We use the normal approximation of the binomial distribution because $N$ becomes sufficiently large as $\dt \to 0$.

\Cref{eq:exploration} shows that when the discretization time scale is infinitesimal, it is impossible for the initial random policy to discover a state that produces a positive reward.

\section{Proof of Proposition 2}
\label{sec:proof2}

Recall that we consider the setting with
$\dt \to 0$, $x \to 0$, $\dt < x$, $\nu \to \infty$ and $f(\mathrm{num}) = 0$ in the \textit{AlertThenOff} environment;
thus the return $R(\tau)$ is given by $\xi$.
We first discuss the optimal policy $\pi_{\theta_f}$ for FiGAR-C.
Its optimal policy for $t$, $\pi^t_{\theta_f}(t|s)$, should produce $t \leq x$
because otherwise it has the risk of ending up with $-\nu$ reward, which is not an optimum.
Therefore, we assume that its (stochastic) duration policy $\pi^t_{\theta_f}(t|s)$,
which is parameterized by $\mu_t$, always produces durations that are less than or equal to $x$.
The whole parameters of FiGAR-C's policy become $\theta_f = [\mu, \mu_t^\top]$,
and $\pi_{\theta_f}$ (which consists of $\pi^{\mathrm{off}}_{\theta_f}$, $\pi^{\mathrm{num}}_{\theta_f}$ and $\pi^{t}_{\theta_f}$)
produces deterministic actions for $\mathrm{off}$ and stochastic actions for $\mathrm{num}$ and $t$.
Now we compute a lower bound for the variance of the PG estimator:
\begin{align}
    &\mkern-18mu \mathrm{tr} \left[ \mathbb{V}_{\tau \sim p_{\theta_f}(\tau)}
    \left[ G_{\theta_f}(\tau) \right] \right]\\
    &= \mathrm{tr} \left[ \mathbb{V}_{\tau \sim p_{\theta_f}(\tau)}
    \left[ \left( \sum_{i=0}^{N-1} \nabla_{\theta_f} \log \pi_{\theta_f} (\mathrm{num}_i, t_i | s_i) \right) R(\tau) \right] \right]\\
    &\geq \mathbb{V}_{\tau \sim p_{\theta_f}(\tau)}
    \left[ \left( \sum_{i=0}^{N-1} \frac{\partial}{\partial \mu} \log \pi^{\mathrm{num}}_{\theta_f} (\mathrm{num}_i | s_i) \right) R(\tau) \right] \label{eq:figar_explosion_ineq}\\
    &= \mathop{\mathbb{V}}_{\substack{\{\epsilon_{i}\} \overset{\text{i.i.d.}}{\sim} \mathcal{N}(0, 1)
    \\ s_{0:N} \sim p_{\theta_f}(s_{0:N}|\{\epsilon_{i}\})}}
    \left[ \left( \sum_{i=0}^{N-1} \epsilon_{i} \right) R(\tau) \right]\\
    &\geq \mathbb{E}_{\{\epsilon_{i}\} \overset{\text{i.i.d.}}{\sim} \mathcal{N}(0, 1)} \left[ 
    \left(\sum_{i=0}^{N-1} \epsilon_{i} \right)^2
    \mathbb{V}_{s_{0:N} \sim p_{\theta_f}(s_{0:N}|\{\epsilon_{i}\})} \left[ R(\tau) \right] \right]\\
    &= \mathbb{E}_{\{\epsilon_{i}\} \overset{\text{i.i.d.}}{\sim} \mathcal{N}(0, 1)} \left[ 
    \left(\sum_{i=0}^{N-1} \epsilon_{i} \right)^2
    \mathbb{V}_{\xi \sim \mathcal{N}(0, 1)} \left[ \xi \right] \right]\\
    &= N \geq \frac{1}{x} \label{eq:figar_explosion},
\end{align}
where $N$ denotes the number of decision steps.
We reparameterize actions as in \Cref{eq:before_tv} and use the law of total variance.
For simplicity, we assume that FiGAR-C's duration policy is stochastic,
but \Cref{eq:figar_explosion} still holds even if the duration policy is deterministic or fixed (in this case, $\theta_f$ becomes $[\mu]$) due to the inequality in \Cref{eq:figar_explosion_ineq}.

From \Cref{eq:figar_explosion}, we can find that when $x \to 0$,
both the variance of the policy gradient and the number of decision steps explode to infinity.

On the other hand, let us consider SAR's optimal policy $\pi_{\theta_s}$.
If we set $\Delta(s_1, s_2) = |s_1 - s_2|$,
one of the optimal (deterministic) policies for $d$ can simply be $\pi^d_{\theta_s}(s) = \frac{1}{2}$.
In this case, the whole parameters of SAR's policy become $\theta_s = [\mu]$,
and $\pi_{\theta_s}$ (which consists of $\pi^{\mathrm{off}}_{\theta_s}$, $\pi^{\mathrm{num}}_{\theta_s}$ and $\pi^{d}_{\theta_s}$)
produces stochastic actions for $\mathrm{num}$ and deterministic actions for $\mathrm{off}$ and $d$.
Also, $N$ becomes $2$, as it stops an action only once when $s$ changes to $1\ (\text{alerted})$.
We can then compute the variance of the PG estimator as follows:
\begin{align}
    &\mkern-18mu \mathrm{tr} \left[\mathbb{V}_{\tau \sim p_{\theta_s}(\tau)}
    \left[ G_{\theta_s}(\tau) \right] \right]\\
    &= \mathbb{V}_{\tau \sim p_{\theta_s}(\tau)}
    \left[ \left( \sum_{i=0}^{N-1} \nabla_{\theta_s} \log \pi^{\mathrm{num}}_{\theta_s} (\mathrm{num}_i | s_i) \right) R(\tau) \right]\\
    &= \mathop{\mathbb{V}}_{\substack{\{\epsilon_{i}\} \overset{\text{i.i.d.}}{\sim} \mathcal{N}(0, 1)
    \\ s_{0:N} \sim p_{\theta_s}(s_{0:N}|\{\epsilon_{i}\})}}
    \left[ \left( \sum_{i=0}^{N-1} \epsilon_{i} \right) R(\tau) \right]\\
    &= \mathbb{V}_{\{\epsilon_{i}\} \overset{\text{i.i.d.}}{\sim} \mathcal{N}(0, 1), \xi \sim \mathcal{N}(0, 1)}
    \left[ \left( \sum_{i=0}^{N-1} \epsilon_{i} \right) \xi \right]\\
    &= \mathbb{V}_{\{\epsilon_{i}\} \overset{\text{i.i.d.}}{\sim} \mathcal{N}(0, 1), \xi \sim \mathcal{N}(0, 1)}
    \left[ \left( \epsilon_0 + \epsilon_1 \right) \xi \right]\\
    &= 2.
\end{align}

Therefore, we conclude that the optimal policy for SAR does not suffer from
either variance explosion or infinite decision steps in \textit{AlertThenOff} environment, even if $x \to 0$.

\clearpage

\section{Additional Results}
\label{sec:add_results}

\begin{figure}[t!]
  \centering
  \includegraphics[width=\linewidth]{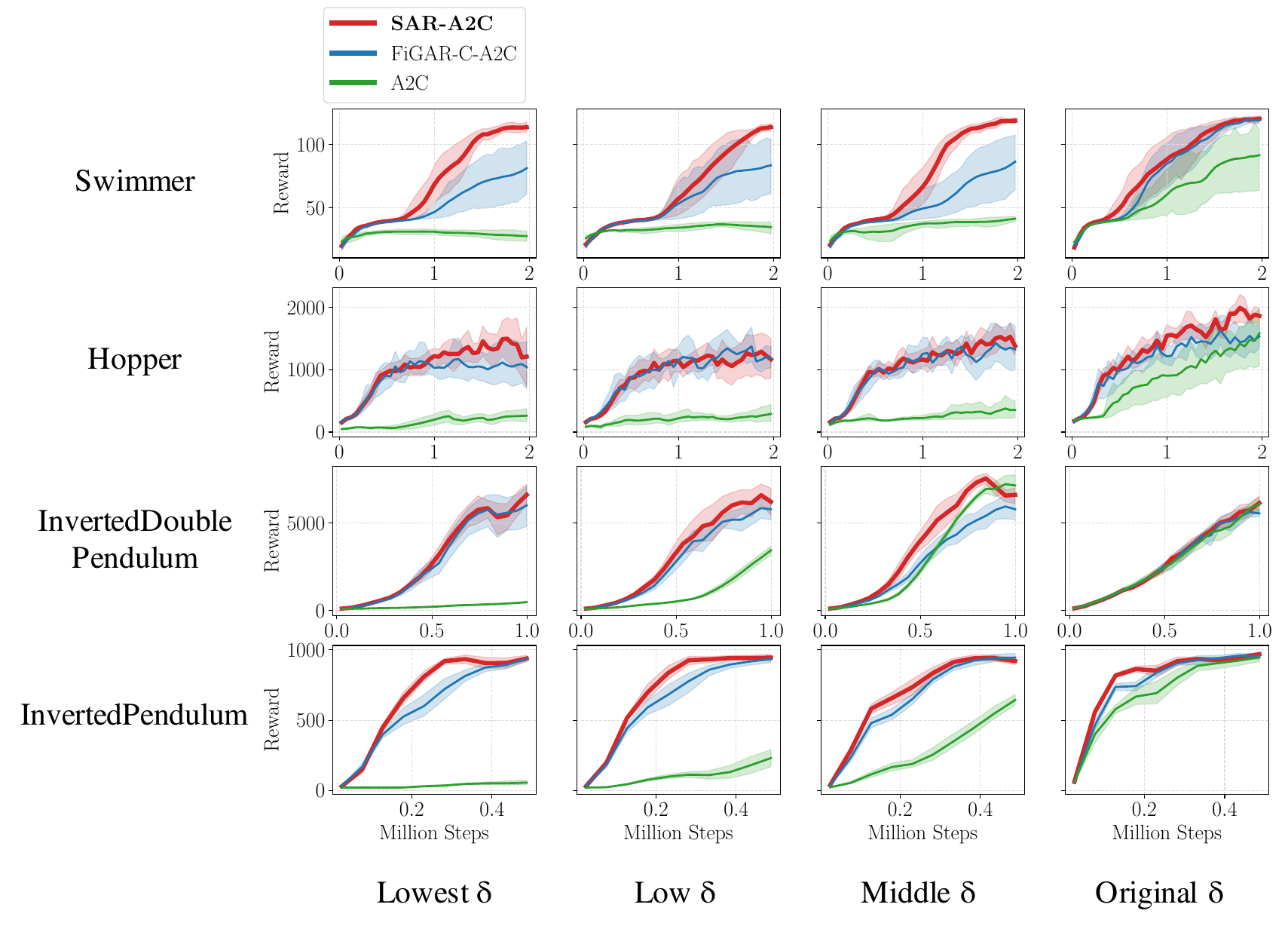}
  \caption{
      Training curves of SAR-A2C, FiGAR-C-A2C and A2C on four deterministic MuJoCo environments with various $\dt$'s.
      Shaded areas represent the $95\%$ confidence intervals over eight runs.
  }
  \label{fig:determ_a2c}
\end{figure}

\textbf{Deterministic Environments.}
We train SAR-A2C, FiGAR-C-A2C, A2C on the four environments of Swimmer-v2, Hopper-v2, InvertedDoublePendulum-v2 and InvertedPendulum-v2,
whose results are shown in \Cref{fig:determ_a2c}.
We find that A2C struggles to perform well on complex environments such as Ant-v2.
SAR mostly shows $\dt$-invariance, outperforming the baselines in most of the environments.

\begin{figure}[t!]
  \centering
  \includegraphics[width=\linewidth]{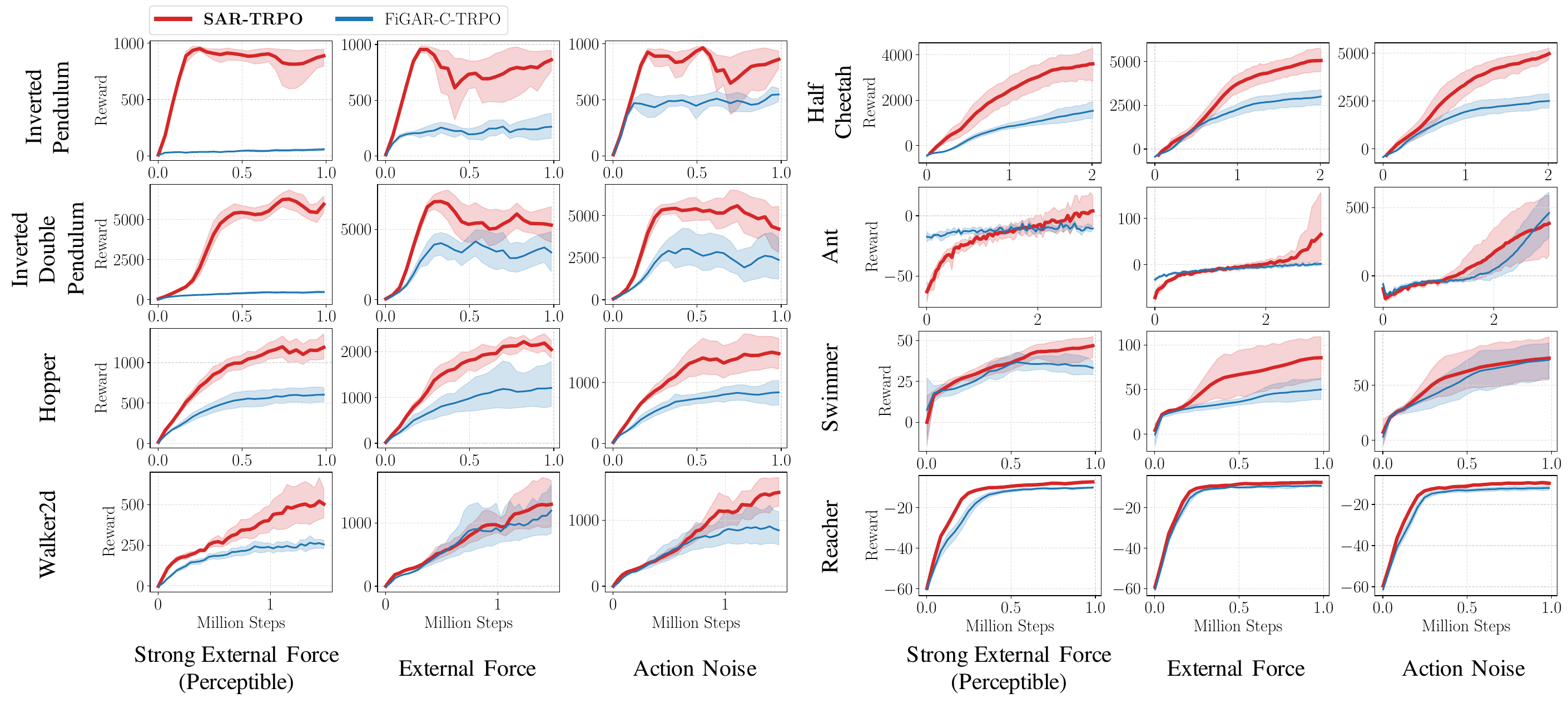}
  \caption{
      Training curves of SAR-TRPO and FiGAR-C-TRPO on eight MuJoCo environments with various types of stochasticity.
      Shaded areas represent the $95\%$ confidence intervals over eight runs.
  }
  \label{fig:stoch_trpo}
\end{figure}

\textbf{Stochastic Environments.}
We also provide the result comparing SAR-TRPO to FiGAR-C-TRPO on eight stochastic MuJoCo environments in \Cref{fig:stoch_trpo}.
As in the case of the PPO baseline, SAR-TRPO mostly demonstrates stronger performance than FiGAR-C-TRPO. 

\begin{figure}[t!]
  \centering
  \includegraphics[width=\linewidth]{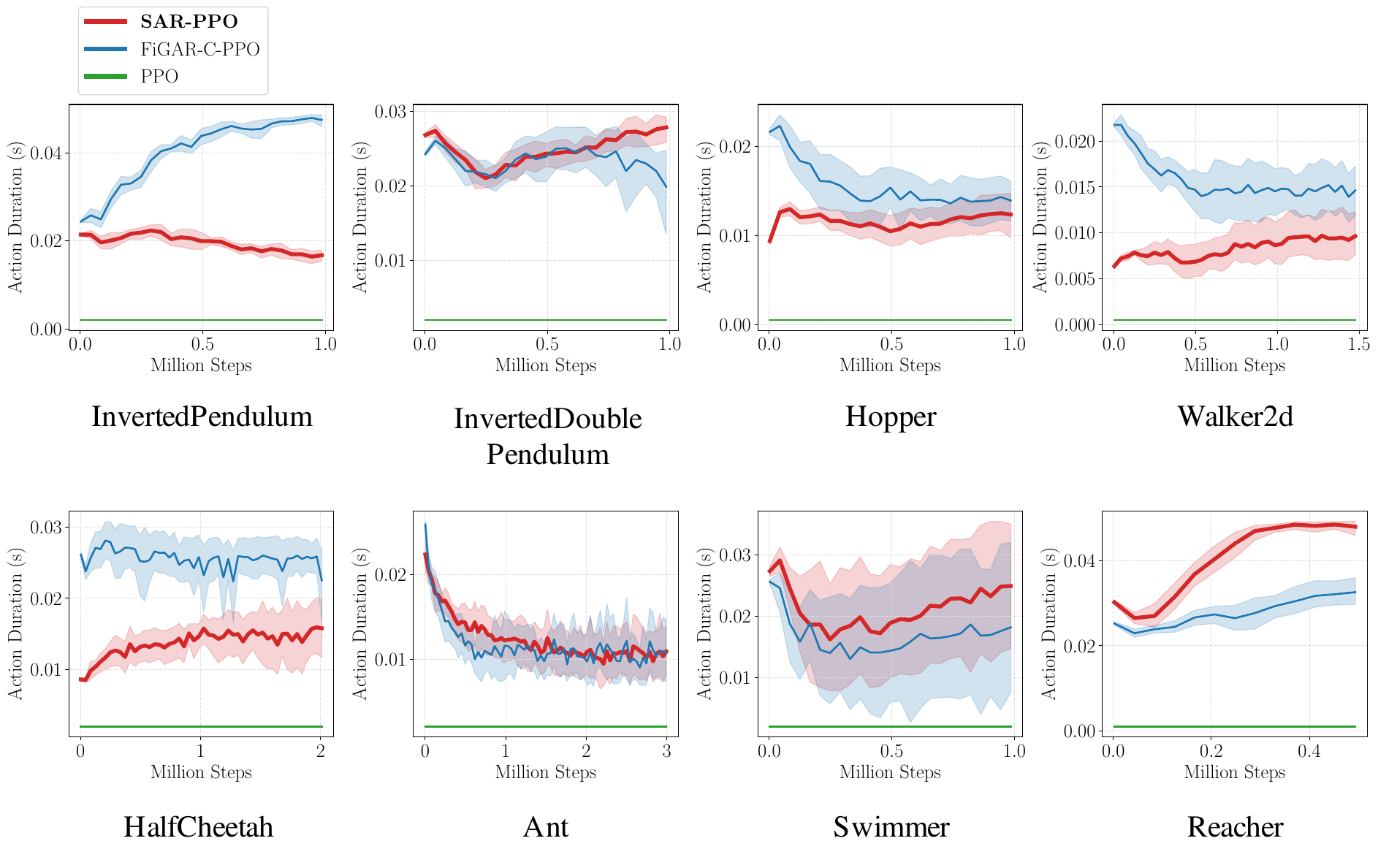}
  \caption{
      Changes in the average action durations of SAR-PPO, FiGAR-C-PPO and PPO on eight deterministic MuJoCo environments with the lowest-$\dt$ settings.
      Shaded areas represent the $95\%$ confidence intervals over eight runs.
  }
  \label{fig:action_dur}
\end{figure}

\textbf{Average action duration.}
We provide how the average action durations of SAR-PPO, FiGAR-C-PPO and PPO change as they are trained.
\Cref{fig:action_dur} demonstrates the results on eight MuJoCo environments with the lowest-$\dt$ settings.

\section{Experiments with Varying Stochasticity Levels}
\label{sec:exp_varying}

\begin{figure}[t!]
  \centering
  \includegraphics[width=\linewidth]{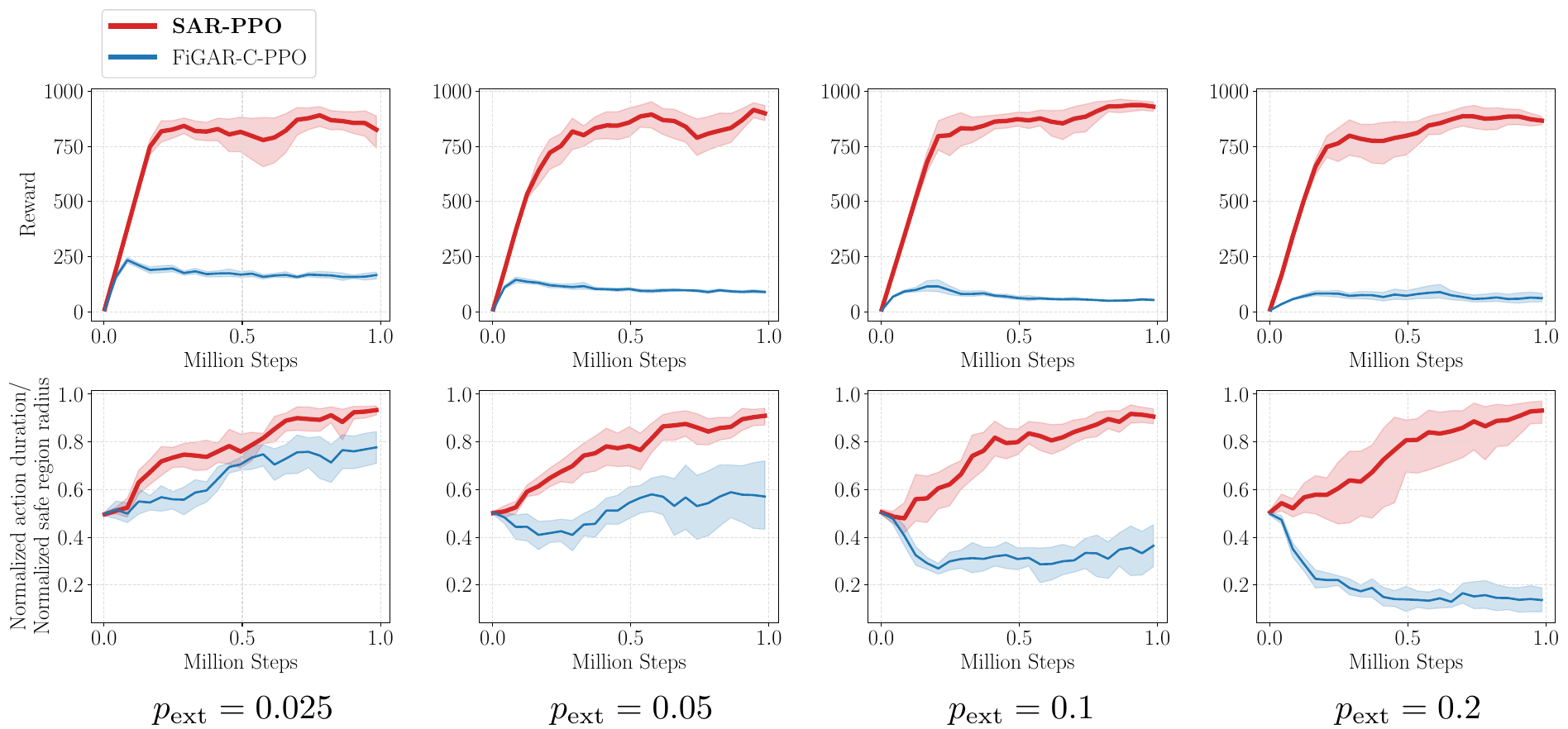}
  \caption{
      Training curves of SAR-PPO and FiGAR-PPO on InvertedPendulum-v2 with the lowest-$\dt$ setting,
      in which the stochasticity level $p_{\text{ext}}$ varies from $0.025$ to $0.2$.
      The first row shows the average performance
      and the second row shows the average normalized action duration (FiGAR-C-PPO) or safe region radius (SAR-PPO).
      Shaded areas represent the $95\%$ confidence intervals over eight runs.
  }
  \label{fig:inp_stoch}
\end{figure}

To further examine how SAR and FiGAR-C evolves as the stochasticity level increases,
we perform an experiment on stochastic InvertedPendulum-v2 ($\dt = 0.002$) with external forces.
We train SAR-PPO and FiGAR-C-PPO on the environment with $p_{\text{ext}} \in \{0.025, 0.05, 0.1, 0.2\}$,
where $p_{\text{ext}}$ denotes the probability of an external force being applied.
\Cref{fig:inp_stoch} shows the plots of the average reward
and the learned (normalized) action duration or safe region radius of each setting.
From the second row, we can observe that the learned action duration decreases as the stochasticity level increases in FiGAR-C,
while such shrinkage does not happen in SAR.
This is because FiGAR-C should reduce action durations when the stochasticity level increases
in order to quickly respond to unexpected events.
Since FiGAR-C is unaware of underlying state changes,
its best strategy is to shorten the duration of actions to be more responsive.
On the other hand, SAR does not necessarily shrink the size of safe regions even if the stochasticity level increases
because it can easily detect the presence of unexpected events by appropriately setting its safe region sizes.
As a result, SAR can handle stochasticity more robustly
as well as preventing the variance explosion problem caused by too short action durations.

\clearpage
\section{Further Demonstrations of PG Methods' Failure with a Low $\dt$}
\label{sec:further_pg}

\subsection{Variance Explosion of the PG Estimator}
\label{sec:var}

\begin{figure}[t!]
  \centering
  \includegraphics[width=0.8\linewidth]{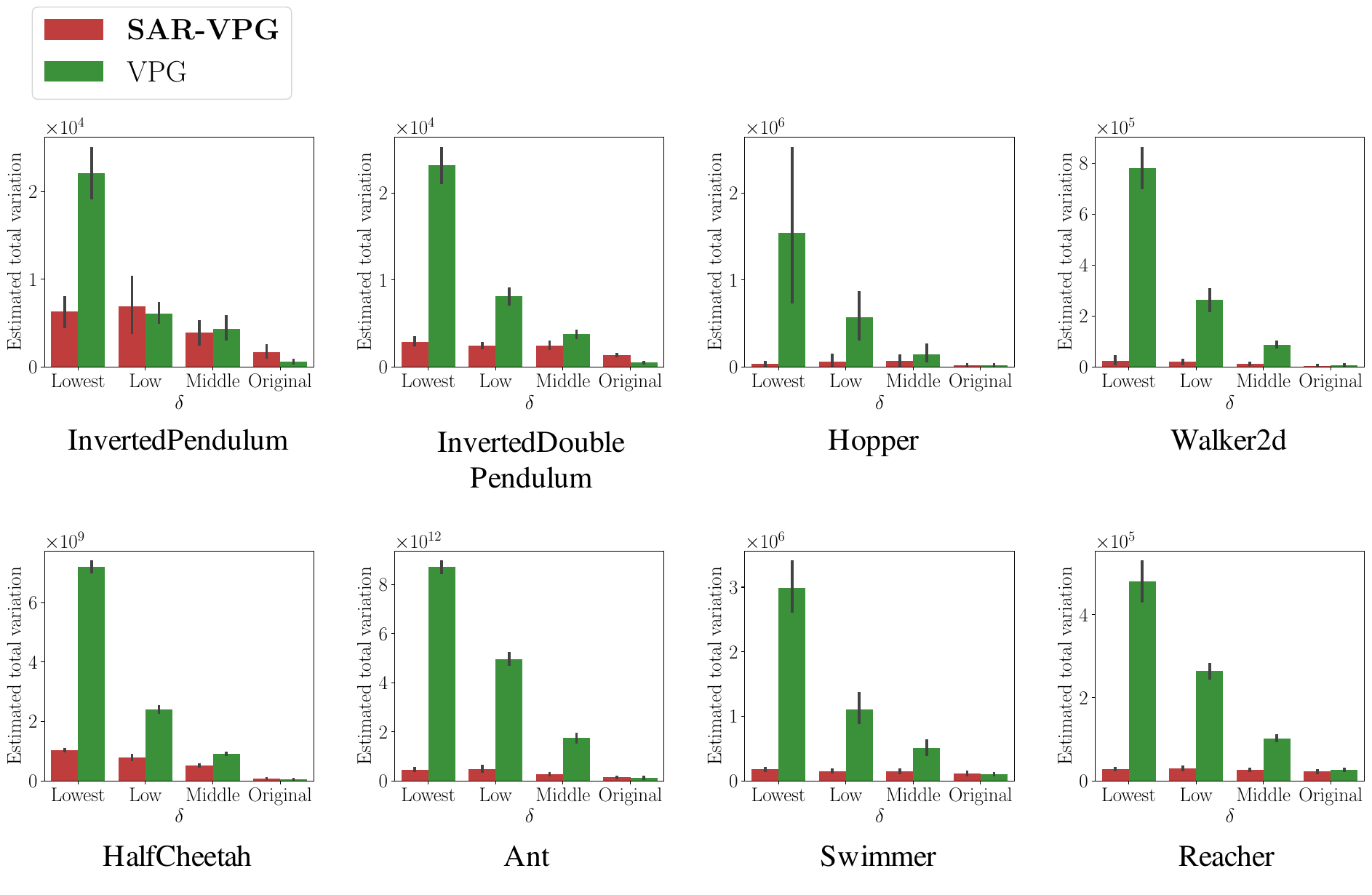}
  \caption{
      Bar plots showing the estimated total variations
      $\mathrm{tr} [ \mathbb{\hat{V}}_{\tau \sim p_\theta(\tau)} [ G_\theta(\tau) ] ]$
      of SAR-VPG and VPG on eight deterministic MuJoCo environments with various $\dt$'s.
      We estimate the total variation with the initial policy.
      Error bars represent the $95\%$ confidence intervals over eight runs.
  }
  \label{fig:var_vpg}
\end{figure}
\begin{figure}[t!]
  \centering
  \includegraphics[width=0.8\linewidth]{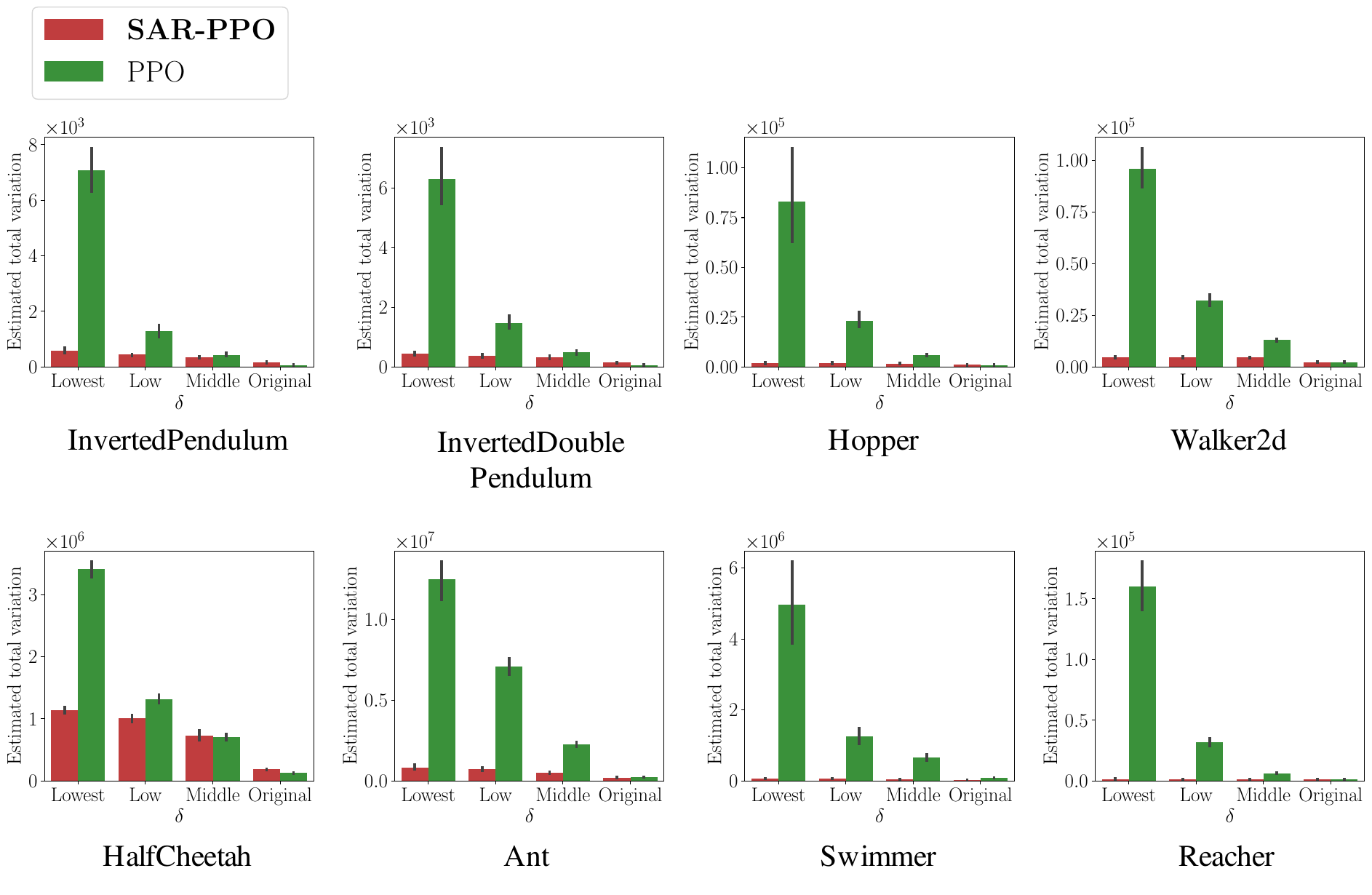}
  \caption{
      Bar plots showing the estimated total variations
      $\mathrm{tr} [ \mathbb{\hat{V}}_{\tau \sim p_\theta(\tau)} [ G_\theta(\tau) ] ]$
      of SAR-PPO and PPO on eight deterministic MuJoCo environments with various $\dt$'s.
      We estimate the total variation with the initial policy.
      Error bars represent the $95\%$ confidence intervals over eight runs.
  }
  \label{fig:var_ppo}
\end{figure}

As shown in \Cref{th:variance},
policy gradient methods are subjected to the variance explosion problem with an exceedingly small $\dt$.
In this section, we empirically demonstrate this phenomenon on eight MuJoCo environments.
We estimate the total variation
$\mathrm{tr} [ \mathbb{\hat{V}}_{\tau \sim p_\theta(\tau)} [ G_\theta(\tau) ] ]$
with the two baseline policy gradient methods: Vanilla Policy Gradient (VPG) and PPO.
In VPG, we do not use any technique for variance reduction such as value functions and reward-to-go policy gradient;
hence, the formula for its gradient estimator is identical to \Cref{eq:g_tau}.
In PPO, we employ the same implementation used for our main results,
including multiple variance reduction techniques such as the GAE \citep{gae_schulman2016}.
We estimate the total variation with $100$ randomly sampled trajectories
(after sampling $10$ trajectories for an initial burn-in phase)
on each of eight randomly initialized policies.
\Cref{fig:var_vpg,fig:var_ppo} show the estimated total variations of both the baseline methods and SAR with various $\dt$'s.
These results confirm that variance explosion empirically occurs on both VPG and PPO with lower-$\dt$ settings, whereas our SAR method can alleviate such a problem.

\subsection{Further Demonstrations of PPO with a Low $\dt$}
\label{sec:further_ppo}

\begin{figure}[t!]
  \centering
  \begin{subfigure}[t]{0.8\linewidth}
      \includegraphics[width=\linewidth]{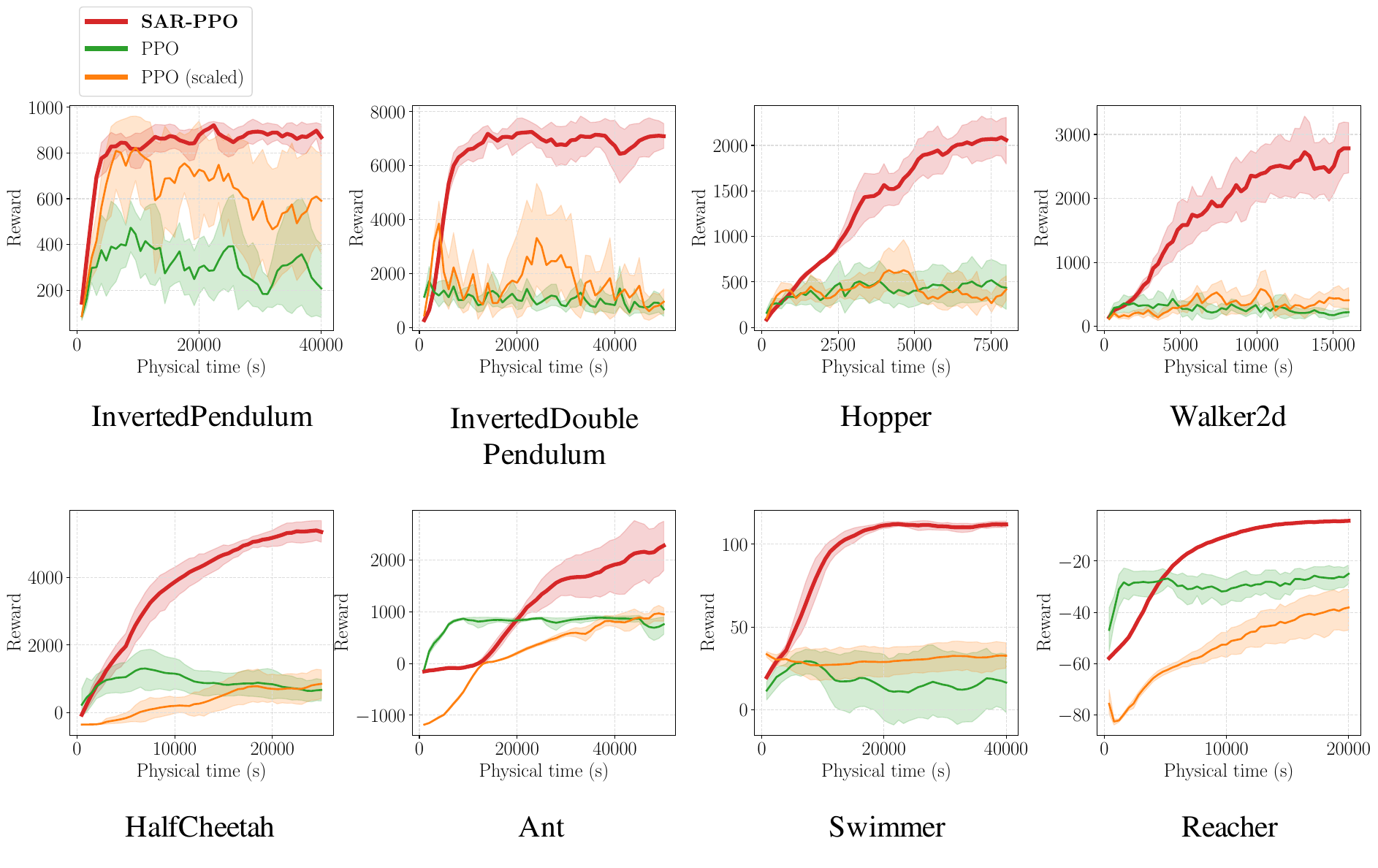}
      \caption{Physical time}
  \end{subfigure}
  \begin{subfigure}[t]{0.8\linewidth}
      \includegraphics[width=\linewidth]{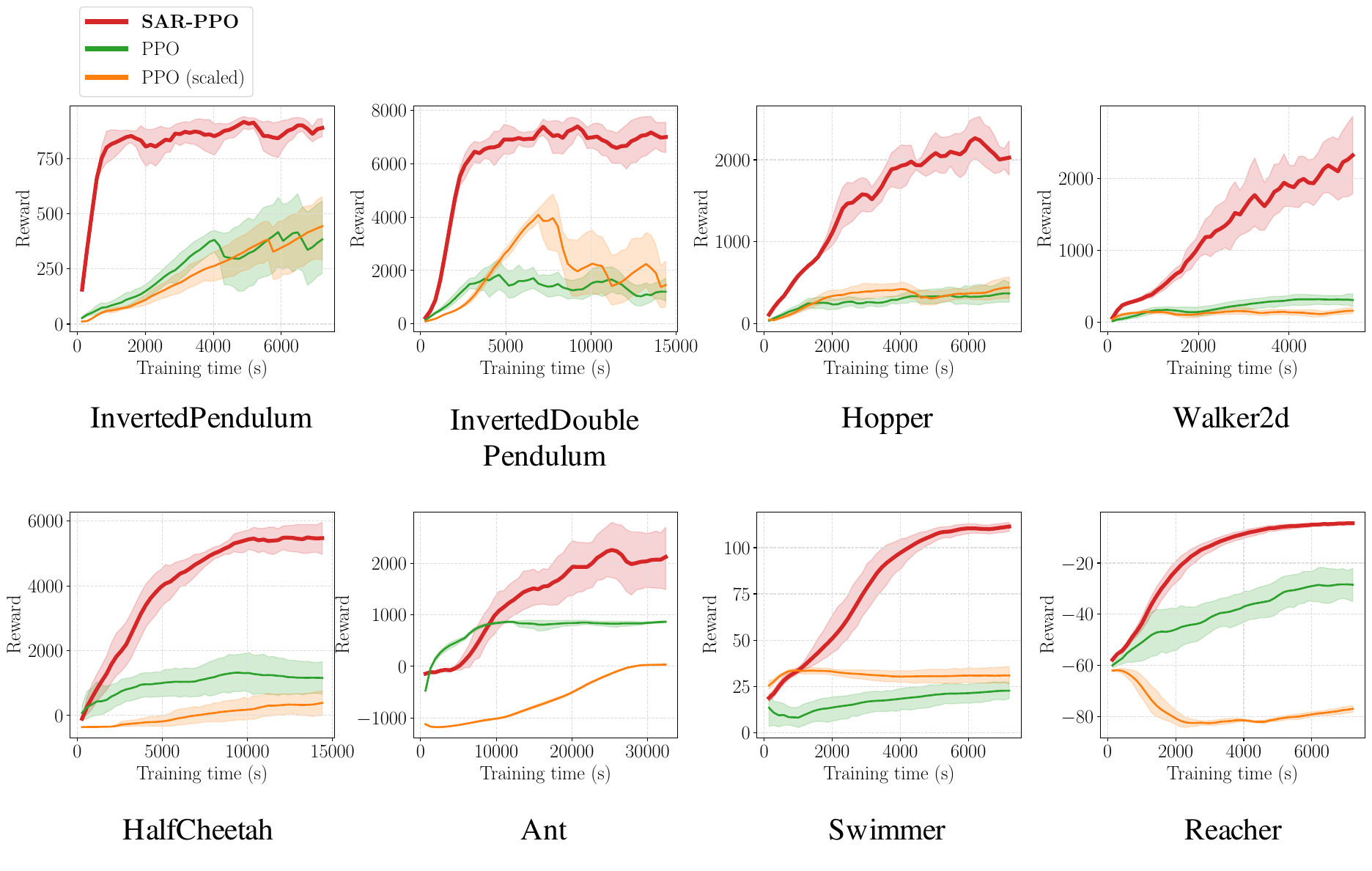}
      \caption{Training time}
  \end{subfigure}
  \caption{
      Training curves of SAR-PPO and PPO's variants with respect to (a) physical time and (b) training time for $x$-axis on eight deterministic MuJoCo environments with the lowest-$\dt$ settings.
      Shaded areas represent the $95\%$ confidence intervals over eight runs.
  }
  \label{fig:ppo_multiple_x}
\end{figure}

We verify both theoretically (\Cref{sec:dt0}) and empirically (\Cref{sec:var}) that
PG methods suffer from variance explosion if the learning rate and minibatch size remain the same.
In this section, we show that PG methods still fail in low-$\dt$ settings even if such parameters are properly scaled,
possibly due to the difficulty of exploration (\Cref{sec:dt0}).
We additionally test a variant of PPO (``PPO (scaled)'') with a learning rate scaled by $\dt / \dt_0$ as in \citet{wawrzynski2015},
where $\dt_0$ denotes the original discretization time scale of each environment.
Also, in order to compare them on multiple criteria,
we plot the results on the two $x$-axes of \textit{physical} time and \textit{training} time,
where physical time indicates the time elapsed in the simulated environment
and training time indicates the time elapsed in the real world for training.
\Cref{fig:ppo_multiple_x} demonstrates the training curves on deterministic MuJoCo environments with the lowest-$\dt$ settings.
We observe that both PPO and the scaled PPO variant struggle with small discretization time scales.
On the contrary, SAR-PPO exhibits strong performance compared to the baseline PPO methods.
Furthermore, as revealed by the comparison between the physical time curve and the training time curve of InvertedPendulum-v2,
the result suggests that our method significantly facilitates training via action repetition.

\section{Comparison with Autoregressive Policies}
\label{sec:arp}

\begin{figure}[t!]
  \centering
  \begin{subfigure}[t]{0.8\linewidth}
      \includegraphics[width=\linewidth]{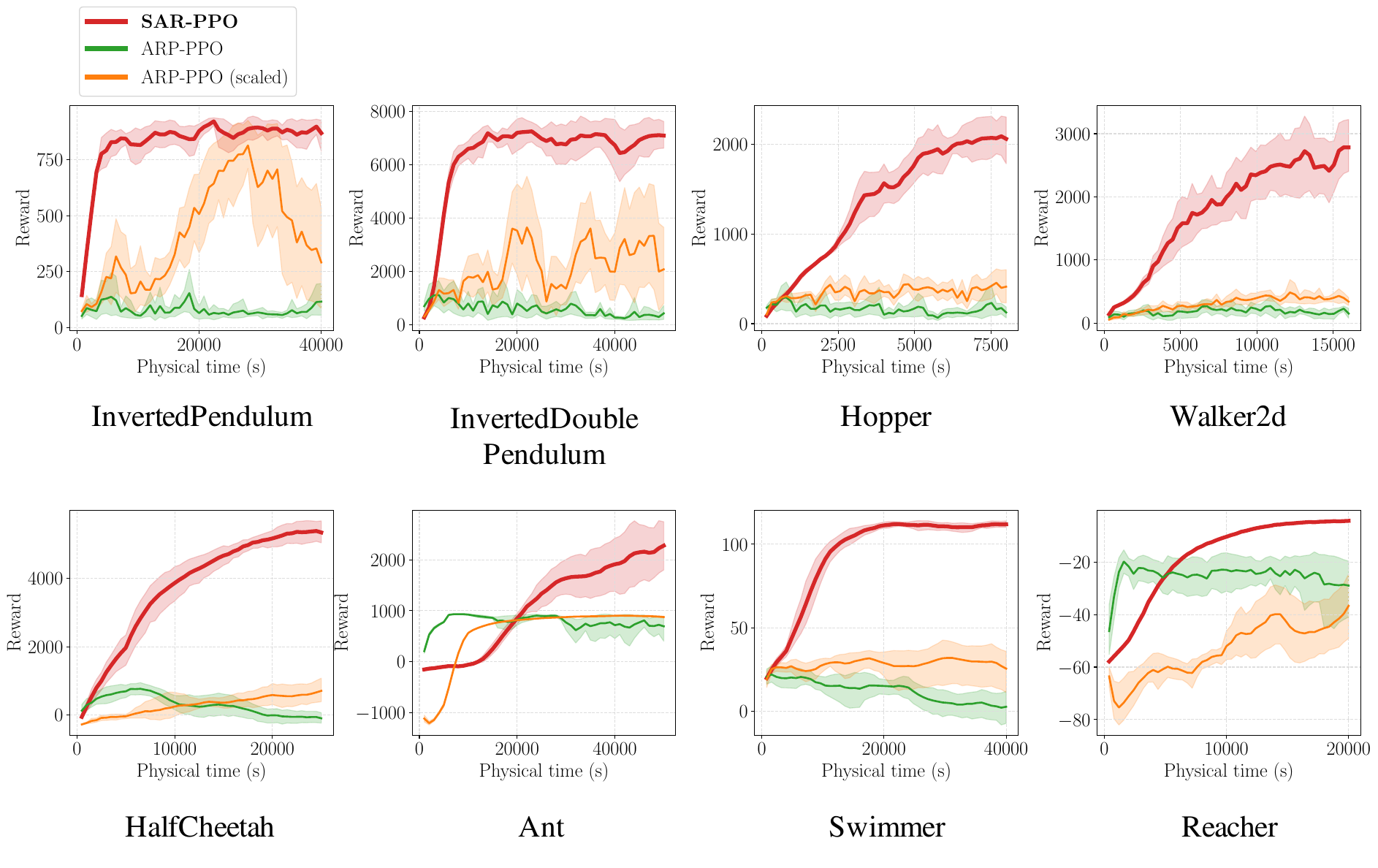}
      \caption{Physical time}
  \end{subfigure}
  \begin{subfigure}[t]{0.8\linewidth}
      \includegraphics[width=\linewidth]{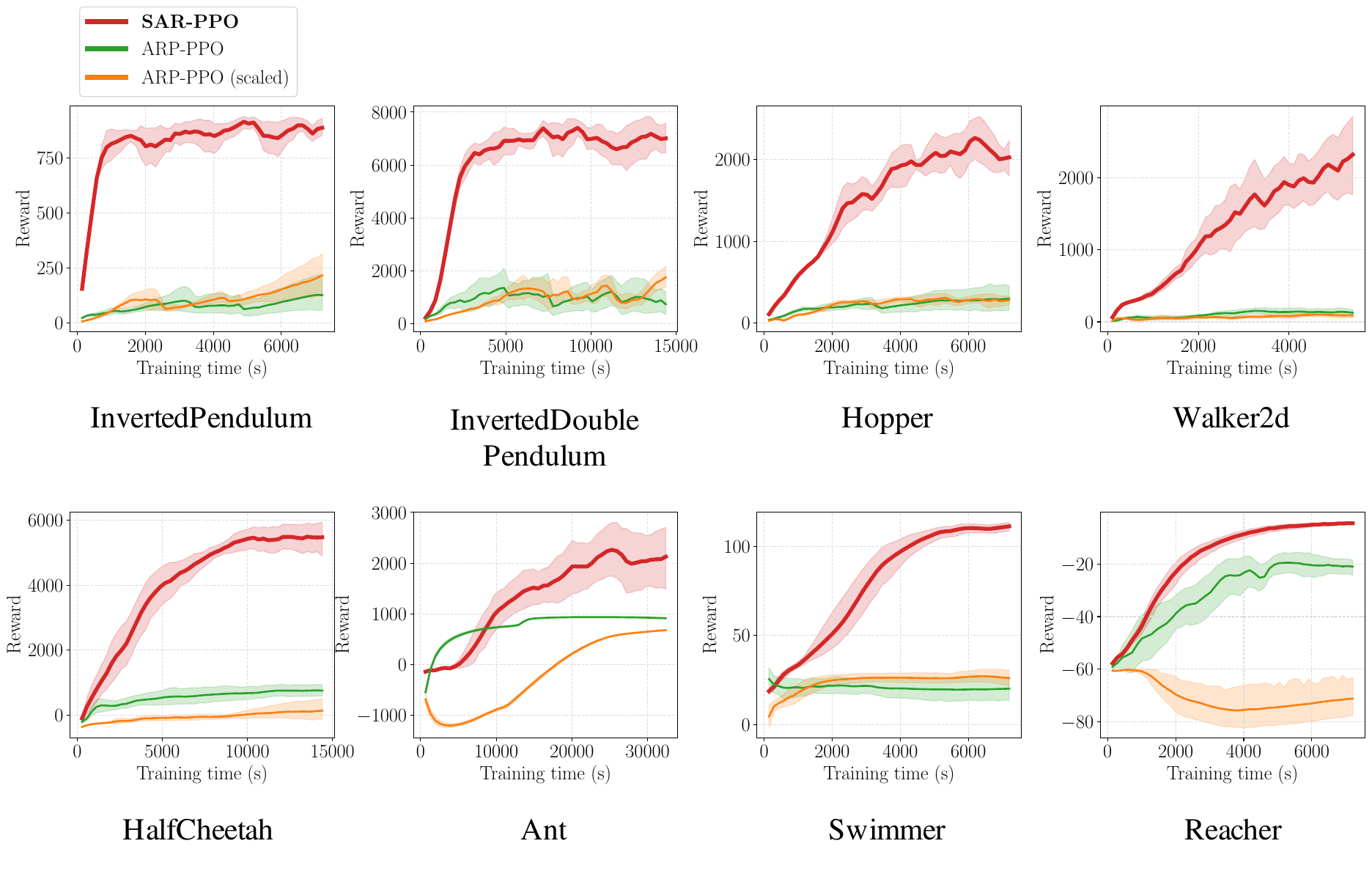}
      \caption{Training time}
  \end{subfigure}
  \caption{
      Training curves of SAR-PPO and ARP-PPO's variants with respect to (a) physical time and (b) training time for $x$-axis on eight deterministic MuJoCo environments with the lowest-$\dt$ settings.
      Shaded areas represent the $95\%$ confidence intervals over eight runs.
  }
  \label{fig:arp}
\end{figure}

We make an additional comparison with autoregressive policies (ARPs) \cite{ar_korenkevych2019},
which use autoregressive processes that could prevent the variance explosion problem with a low $\dt$.
An ARP uses the autoregressive noise process to sample actions so that the actions can be temporally correlated.
It has two main hyperparameters: $p_{\text{ord}}$ and $\alpha$,
where $p_{\text{ord}}$ is the order of the autoregressive process
and $\alpha$ controls its temporal smoothness ($\alpha = 0$ corresponds to the white Gaussian noise).

We compare SAR-PPO to ARPs trained with PPO (``ARP-PPO'')
as well as its variant (``ARP-PPO (scaled)'') with the scaled learning rate specified in \Cref{sec:further_ppo}.
For ARP-PPO and ARP-PPO (scaled),
we respectively perform hyperparameter search over $p \in \{1, 3\}$ and $\alpha \in \{0.3, 0.5, 0.8, 0.95\}$.
We individually tune the hyperparameters on each environment,
while we share the hyperparameter $d_{\tmax} = 0.5$ across all the environments in the case of SAR.
The hyperparameters used for ARPs are as follows:
\begin{itemize}
    \item Ant-v2: $p = 3$, $\alpha = 0.5$ for ARP-PPO and $p = 3$, $\alpha = 0.5$ for ARP-PPO (scaled).
    \item HalfCheetah-v2: $p = 3$, $\alpha = 0.3$ for ARP-PPO and $p = 1$, $\alpha = 0.8$ for ARP-PPO (scaled).
    \item InvertedDoublePendulum-v2: $p = 1$, $\alpha = 0.5$ for ARP-PPO and $p = 1$, $\alpha = 0.95$ for ARP-PPO (scaled).
    \item InvertedPendulum-v2: $p = 1$, $\alpha = 0.3$ for ARP-PPO and $p = 1$, $\alpha = 0.95$ for ARP-PPO (scaled).
    \item Swimmer-v2: $p = 1$, $\alpha = 0.95$ for ARP-PPO and $p = 3$, $\alpha = 0.8$ for ARP-PPO (scaled).
    \item Reacher-v2: $p = 1$, $\alpha = 0.3$ for ARP-PPO and $p = 3$, $\alpha = 0.8$ for ARP-PPO (scaled).
    \item Hopper-v2: $p = 3$, $\alpha = 0.5$ for ARP-PPO and $p = 1$, $\alpha = 0.95$ for ARP-PPO (scaled).
    \item Walker2d-v2: $p = 1$, $\alpha = 0.5$ for ARP-PPO and $p = 1$, $\alpha = 0.8$ for ARP-PPO (scaled).
\end{itemize}

\Cref{fig:arp} shows the training curves with respect to both physical time and training time (details in \Cref{sec:further_ppo}).
It is observed that SAR outperforms ARPs often by a large margin on both criteria.

\section{Results with $\lambda$-SAR}
\label{sec:lsar}

\begin{figure}[t!]
  \centering
  \includegraphics[width=\linewidth]{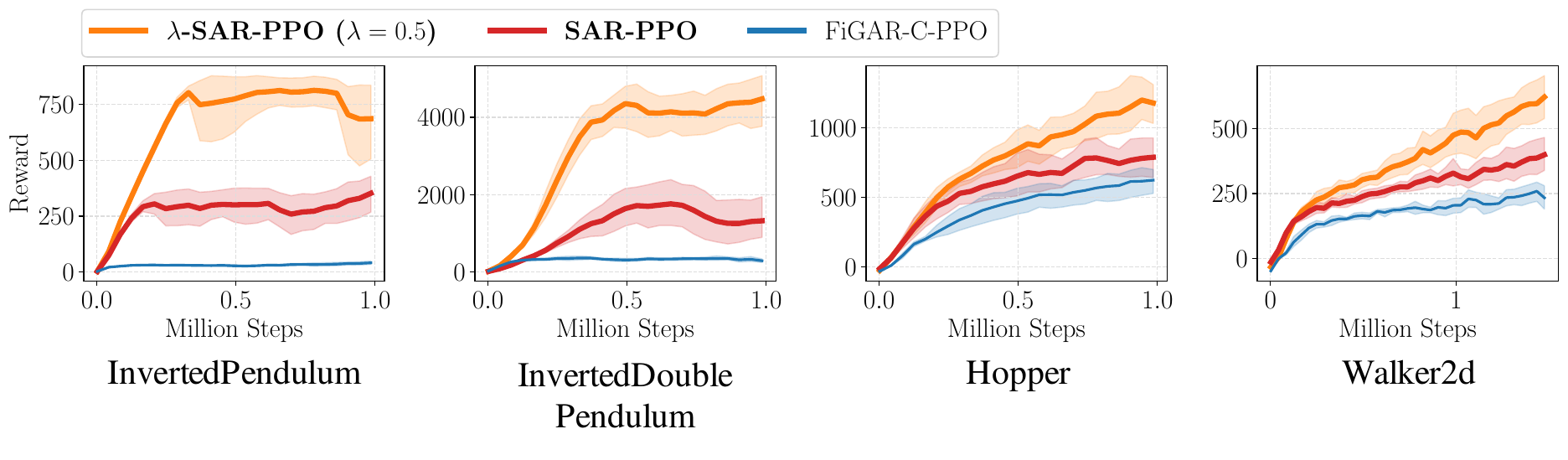}
  \caption{
      Training curves of $\lambda$-SAR-PPO ($\lambda = 0.5$), SAR-PPO and FiGAR-C-PPO
      on four stochastic POMDP environments with the lowest-$\dt$ settings.
      Shaded areas represent the $95\%$ confidence intervals over eight runs.
      $\lambda$-SAR shows better performance compared to the others in these POMDP settings.
  }
  \label{fig:lsar}
\end{figure}

To verify whether $\lambda$-SAR described in \Cref{sec:sar_nme} could be effective in partially observable MDPs (POMDPs),
we modify the stochastic MuJoCo environments with the ``Strong External Force (Perceptible)'' setting (described in \Cref{sec:exp_stoch})
by adding partial observability.
Specifically, in addition to the stochasticity, we make the environment cause a penalty reward of $r_{\text{penalty}}$
when the agent holds the same action more than or equal to $t_{\text{thres}}$ seconds,
where the agent cannot observe the current holding time of an action,
which renders the environment to be a POMDP.
We test methods on the four MuJoCo environments of InvertedPendulum-v2, InvertedDoublePendulum-v2, Hopper-v2 and Walker2d-v2 with the lowest $\dt$'s,
where we set $t_{\text{thres}} = 0.04$ and $r_{\text{penalty}} = -1$ for InvertedPendulum-v2,
$t_{\text{thres}} = 0.04$ and $r_{\text{penalty}} = -10$ for InvertedDoublePendulum-v2,
and $t_{\text{thres}} = 0.025$ and $r_{\text{penalty}} = -20$ for Hopper-v2 and Walker2d-v2.
\Cref{fig:lsar} shows the training curves of $\lambda$-SAR-PPO ($\lambda = 0.5$), SAR-PPO and FiGAR-C-PPO.
We confirm that $\lambda$-SAR can cope with such partial observability by incorporating temporal information into safe regions.

\section{Ablation Study}
\label{sec:abl}

\begin{figure}[t!]
  \centering
  \includegraphics[width=\linewidth]{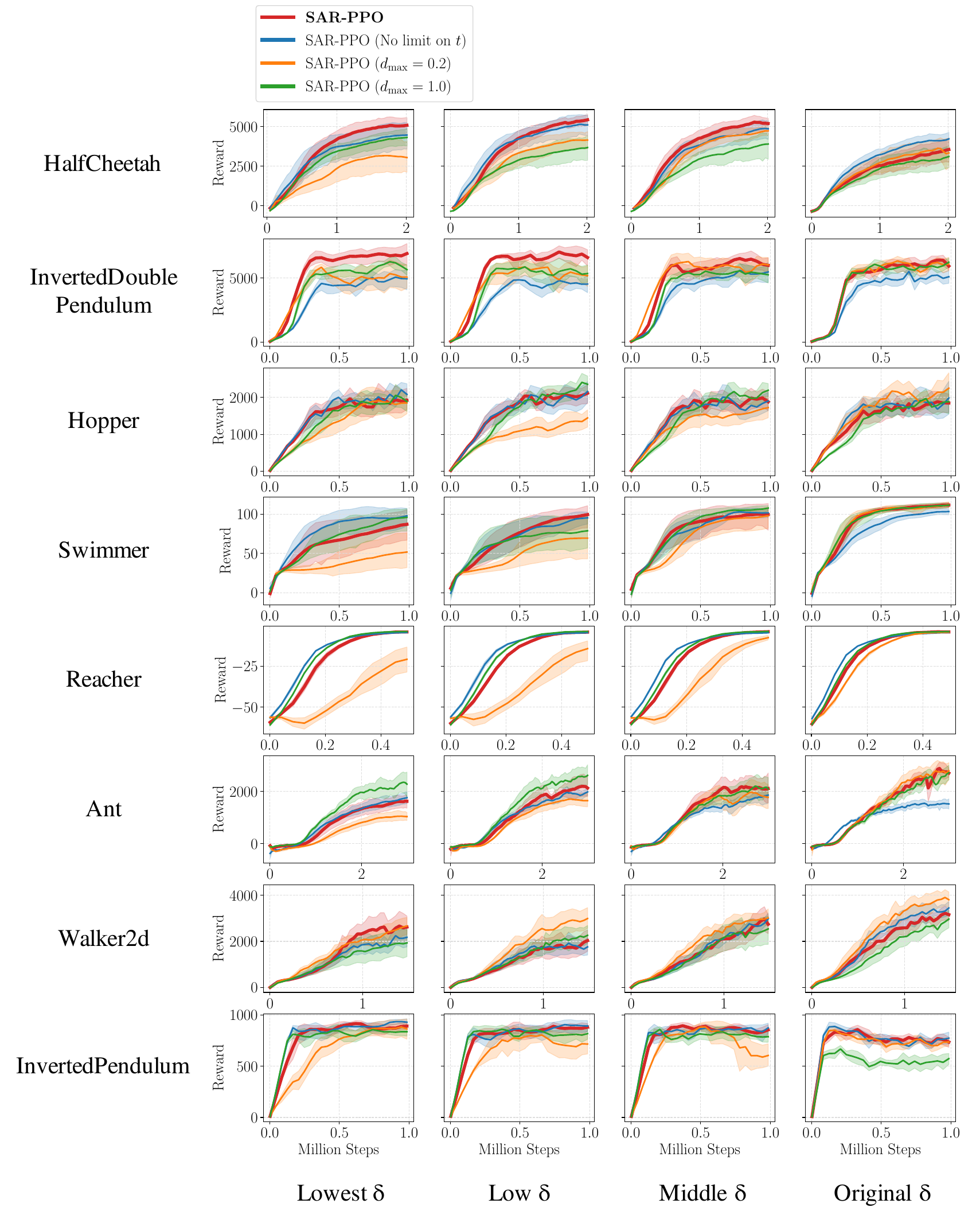}
  \caption{
      Training curves of multiple variations of SAR-PPO on eight deterministic MuJoCo environments with various $\dt$'s.
      Shaded areas represent the $95\%$ confidence intervals over eight runs.
  }
  \label{fig:ppo_ablation}
\end{figure}

\textbf{Variants of SAR-PPO.}
We test SAR-PPO with its variations.
As stated in \Cref{sec:experimental_details}, we fix $d_{\tmax}=0.5$ in SAR for the experiments in the main paper.
We alter $d_{\tmax}$ to $0.2$ (``SAR-PPO ($d_{\tmax} = 0.2$)'') or $1.0$ (``SAR-PPO ($d_{\tmax} = 1.0$)'')
to demonstrate how this hyperparameter affects the performance of SAR.
We also experiment with another variant of SAR (``SAR-PPO (No limit on $t$)'')
that does not impose an upper limit on the maximum duration of actions.
\Cref{fig:ppo_ablation} shows the results on eight deterministic MuJoCo environments.
We observe that a small $d_{\tmax}$ may lead to inferior performance on average
since it may excessively limit action durations,
increasing the average number of decision steps and thus the variance of the PG estimator.

\begin{figure}[t!]
  \centering
  \includegraphics[width=\linewidth]{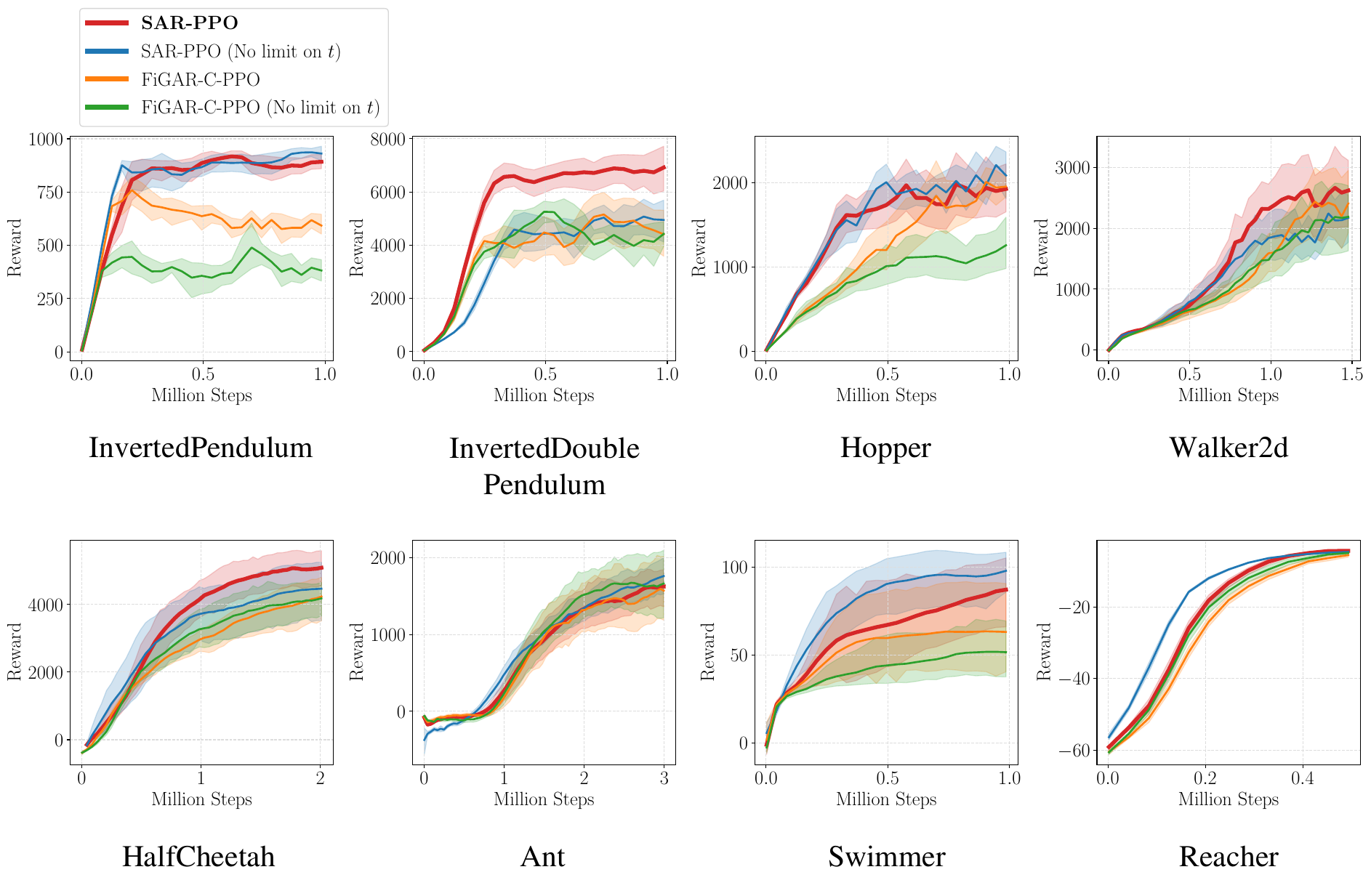}
  \caption{
      Training curves of SAR-PPO, FiGAR-C-PPO and their variants without $t$ limit
      on eight deterministic MuJoCo environments with the lowest-$\dt$ settings.
      Shaded areas represent the $95\%$ confidence intervals over eight runs.
  }
  \label{fig:ppo_t_ablation}
\end{figure}

\textbf{Effect of a limit on $t$.}
In order to examine the effect of imposing an upper limit on action durations,
we test variants of SAR-PPO and FiGAR-C-PPO.
``SAR-PPO (No limit on $t$)'' denotes the same setting as the previous experiment
and ``FiGAR-C-PPO (No limit on $t$)'' denotes the setting of FiGAR-C-PPO without clipping $t$
while it uses the same scale of $t$ as our original FiGAR-C-PPO.
\Cref{fig:ppo_t_ablation} demonstrates that in both setting,
imposing a limit on $t$ leads to better performance on most of the environments as it helps stabilize training,
although SAR-PPO (No limit on $t$) sometimes outperforms the original SAR-PPO on some environments such as Reacher-v2.

\begin{figure}[t!]
  \centering
  \includegraphics[width=\linewidth]{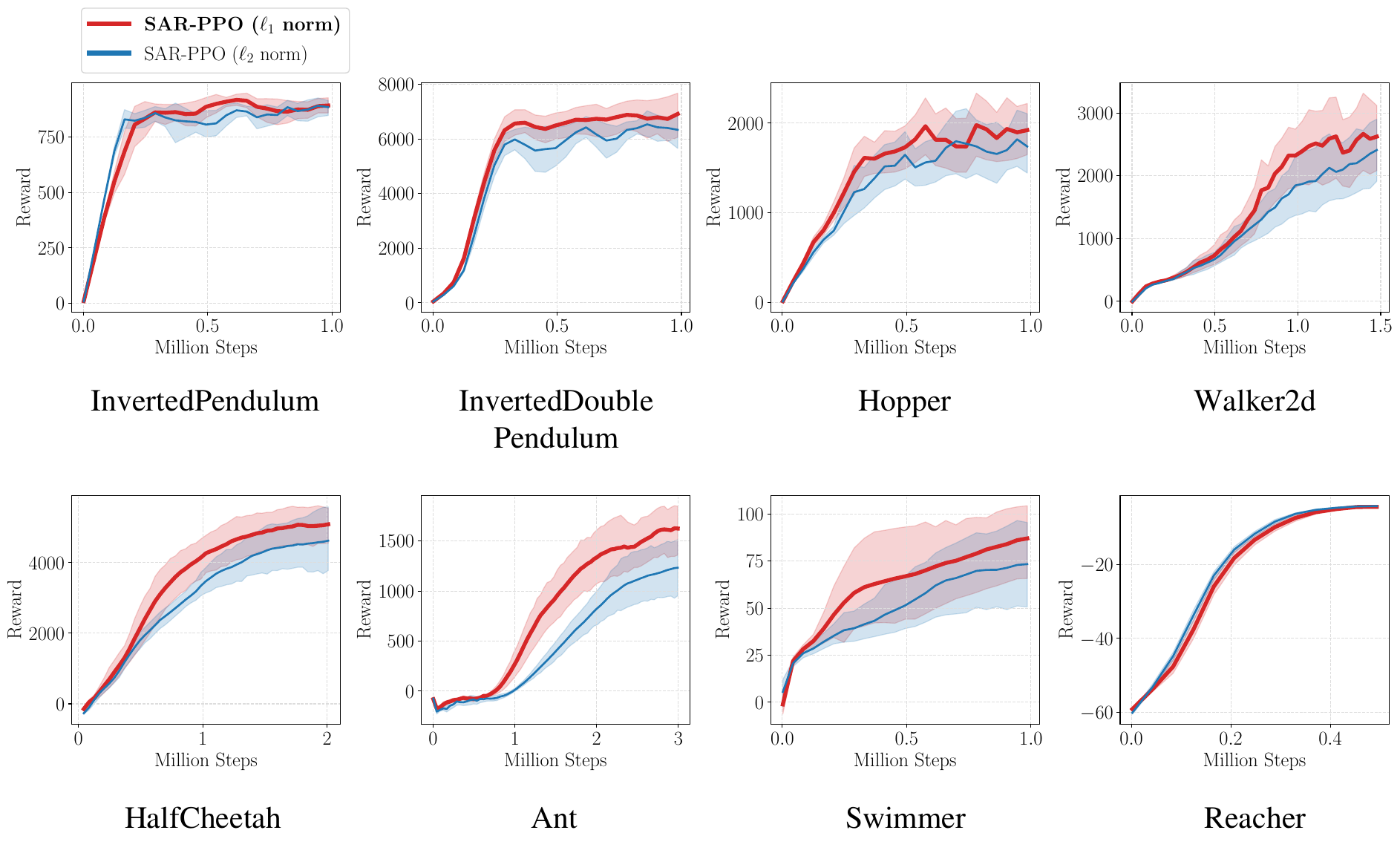}
  \caption{
      Training curves of SAR-PPO with the $\ell_1$ or $\ell_2$ norm
      on eight deterministic MuJoCo environments with the lowest-$\dt$ settings.
      Shaded areas represent the $95\%$ confidence intervals over eight runs.
  }
  \label{fig:ppo_l1_l2}
\end{figure}

\textbf{Variants of SAR-PPO's distance function.}
We use the $\ell_1$ norm for the distance function of SAR:
$\Delta(s, s_i) = \| \tilde{s} - \tilde{s_i} \|_1 / \mathrm{dim}(\mathcal{S})$.
In this experiment, we test another variant of SAR with the $\ell_2$ norm,
whose distance function is defined as $\Delta(s, s_i) = \| \tilde{s} - \tilde{s_i} \|_2 / \sqrt{\mathrm{dim}(\mathcal{S})}$.
We set $d_{\tmax} = 0.5$ for the $\ell_1$ norm and $d_{\tmax} = 1.0$ for the $\ell_2$ norm.
\Cref{fig:ppo_l1_l2} suggests that the $\ell_1$ norm is slightly more effective than the $\ell_2$ norm.
We speculate that this is because some state dimensions with large changes may dominate $\ell_2$ distances.

\clearpage

\section{Experimental Details}
\label{sec:experimental_details}

\subsection{Implementation}

We implement SAR and the baseline methods based on the open-source implementations of
Stable Baselines3 \cite{sb3_raffin2019} (a port of Stable Baselines \cite{sb_hill2018} for PyTorch\cite{pytorch_paszke2019}) for PPO \cite{ppo_schulman2017} and A2C \cite{a2c_mnih2016},
and Stable Baselines \cite{sb_hill2018} for TRPO \cite{trpo_schulman2015}.
We use the publicly released official implementations
for DAU \cite{dau_tallec2019} (\url{https://github.com/ctallec/continuous-rl}) 
and ARP \cite{ar_korenkevych2019} (\url{https://github.com/kindredresearch/arp}).
We provide the implementation for our experiments (including licenses) in the anonymous repository at \url{https://vision.snu.ac.kr/projects/sar}.

\subsection{Environments}
We experiment on eight continuous control environments from MuJoCo \cite{mujoco_todorov2012}:
InvertedPendulum-v2, InvertedDoublePendulum-v2, Hopper-v2, Walker2d-v2,
HalfCheetah-v2, Ant-v2, Reacher-v2 and Swimmer-v2.

The environment parameters used in our experiments are as follows:
\begin{itemize}
    \item Ant-v2:
        $\mathcal{S} = \mathbb{R}^{111}$, $\mathcal{A} = [-1, 1]^8$, $\sigma_{\text{act}} = 1$,
        $p_\text{act} = p_\text{ext} = 0.05$, $\sigma_{\text{ext}} = 100$, $\sigma_{\text{ext2}} = 300$.
    \item HalfCheetah-v2:
        $\mathcal{S} = \mathbb{R}^{17}$, $\mathcal{A} = [-1, 1]^6$, $\sigma_{\text{act}} = 1$,
        $p_\text{act} = p_\text{ext} = 0.05$, $\sigma_{\text{ext}} = 30$, $\sigma_{\text{ext2}} = 300$.
    \item InvertedDoublePendulum-v2:
        $\mathcal{S} = \mathbb{R}^{11}$, $\mathcal{A} = [-1, 1]^1$, $\sigma_{\text{act}} = 1$,
        $p_\text{act} = p_\text{ext} = 0.05$, $\sigma_{\text{ext}} = 100$, $\sigma_{\text{ext2}} = 1000$.
    \item InvertedPendulum-v2:
        $\mathcal{S} = \mathbb{R}^{4}$, $\mathcal{A} = [-3, 3]^1$, $\sigma_{\text{act}} = 3$,
        $p_\text{act} = p_\text{ext} = 0.05$, $\sigma_{\text{ext}} = 300$, $\sigma_{\text{ext2}} = 1000$.
    \item Swimmer-v2:
        $\mathcal{S} = \mathbb{R}^{8}$, $\mathcal{A} = [-1, 1]^2$, $\sigma_{\text{act}} = 1$,
        $p_\text{act} = p_\text{ext} = 0.05$, $\sigma_{\text{ext}} = 100$, $\sigma_{\text{ext2}} = 1000$.
    \item Reacher-v2:
        $\mathcal{S} = \mathbb{R}^{11}$, $\mathcal{A} = [-1, 1]^2$, $\sigma_{\text{act}} = 1$,
        $p_\text{act} = p_\text{ext} = 0.05$, $\sigma_{\text{ext}} = 300$, $\sigma_{\text{ext2}} = 1000$.
        Due to its unique environment dynamics, we apply external torques instead of forces.
    \item Hopper-v2:
        $\mathcal{S} = \mathbb{R}^{11}$, $\mathcal{A} = [-1, 1]^3$, $\sigma_{\text{act}} = 1$,
        $p_\text{act} = p_\text{ext} = 0.05$, $\sigma_{\text{ext}} = 30$, $\sigma_{\text{ext2}} = 300$.
    \item Walker2d-v2:
        $\mathcal{S} = \mathbb{R}^{17}$, $\mathcal{A} = [-1, 1]^6$, $\sigma_{\text{act}} = 1$,
        $p_\text{act} = p_\text{ext} = 0.05$, $\sigma_{\text{ext}} = 100$, $\sigma_{\text{ext2}} = 1000$.
\end{itemize}

\begin{table}[t!]
  \caption{Discretization time scales.}
  \label{table:dts}
  \centering
  \begin{tabular}{lllll}
      \toprule
      Environment & $\dt_{\text{lowest}}$ & $\dt_{\text{low}}$ & $\dt_{\text{middle}}$ & $\dt_{\text{original}}$ ($\dt_0$)\\
      \midrule
      Ant-v2 & $2e-3$ & $5e-3$ & $1e-2$ & $5e-2$ \\
      HalfCheetah-v2 & $2e-3$ & $5e-3$ & $1e-2$ & $5e-2$ \\
      InvertedDoublePendulum-v2 & $2e-3$ & $5e-3$ & $1e-2$ & $5e-2$ \\
      InvertedPendulum-v2 & $2e-3$ & $5e-3$ & $1e-2$ & $4e-2$ \\
      Swimmer-v2 & $2e-3$ & $5e-3$ & $1e-2$ & $4e-2$ \\
      Reacher-v2 & $1e-3$ & $2e-3$ & $5e-3$ & $2e-2$ \\
      Hopper-v2 & $5e-4$ & $1e-3$ & $2e-3$ & $8e-3$ \\
      Walker2d-v2 & $5e-4$ & $1e-3$ & $2e-3$ & $8e-3$ \\
      \bottomrule
  \end{tabular}
\end{table}

For the discretization time scales, we generally follow the values from \citet{dau_tallec2019}.
\Cref{table:dts} shows the values we used for $\dt$'s.
We use an episode horizon of $1000$
and a discount factor of $\gamma_0 = 0.99$
for all the environments with the original discretization time scale ($\dt_0$).
In lower-$\dt$ settings, we scale the episode length by $\dt_0 / \dt$ to maintain the same physical time limit,
and set the discount factor to $\gamma_0^{\dt / \dt_0}$ to have the same effective horizon.
We discount the reward both between decision steps (accordingly to \Cref{eq:discount})
and during action repetitions in SAR and FiGAR-C.
For the ``Strong External Force (Perceptible)'' setting described in \Cref{sec:exp_stoch},
we append to the state the applied 3-D force (or 3-D torque in the case of Reacher-v2) vector clipped to a range of $[-1, 1]$.

\subsection{Training}

Throughout the experiments, we model each learnable component with an MLP with two hidden layers of 256 dimensions.
For the policies of PPO, TRPO and A2C,
we use a normal distribution with a learnable diagonal covariance matrix that is independent of states,
following the implementations of \citet{trpo_schulman2015,ppo_schulman2017}.
For the policies of SAR and FiGAR-C,
we modify the variance corresponding to $d$ or $t$ actions to be dependent on states.
We normalize returns (rewards) and each dimension of states using their moving averages
for the inputs of the components in all environments except Ant-v2;
we find that it performs better not to use the normalization in Ant-v2.
For SAR's distance function, we use normalized states in all environments
in order to make safe regions agnostic to the scale of each state dimension.
We set $d_{\tmax} = 0.5$ (chosen among $\{0.1, 0.2, 0.5, 1.0\}$) for SAR
and $t_{\tmax} = 0.05$ (chosen among $\{0.01, 0.02, 0.05, 0.1\}$) for FiGAR-C,
and share them across all the environments.
We run our experiments on our internal CPU cluster mostly consisting of
Intel Xeon E5-2695 v4 and Intel Xeon Gold 6130
processors.
Each run in our experiments usually takes 3-12 hours on a single CPU core.

We report the hyperparameters used for each RL algorithm in \Cref{table:hp_ppo,table:hp_trpo,table:hp_a2c}.
For further implementation details,
we refer to our released code as well as the official implementations of DAU and ARP.

\begin{table}[t!]
  \caption{Hyperparameters for PPO.}
  \label{table:hp_ppo}
  \centering
  \begin{tabular}{ll}
      \toprule
      Hyperparameter & Value \\
      \midrule
      Optimizer & Adam \\
      Learning rate & $1e-4$ \\
      Nonlinearity & ReLU \\
      \# decision steps per train step & $2048$ \\
      \# epochs per train step & $10$ \\
      Minibatch size & $64$ \\
      GAE parameter $\lambda$ & $0.95$ \\
      Clipping parameter $\epsilon$ & $0.2$ \\
      \bottomrule
  \end{tabular}
\end{table}

\begin{table}[t!]
  \caption{Hyperparameters for TRPO.}
  \label{table:hp_trpo}
  \centering
  \begin{tabular}{ll}
      \toprule
      Hyperparameter & Value \\
      \midrule
      Optimizer & Adam \\
      Learning rate & $1e-4$ \\
      Nonlinearity & ReLU \\
      \# decision steps per train step & $1024$ \\
      GAE parameter $\lambda$ & $0.95$ \\
      KL step size & $0.01$ \\
      Conjugate gradient damping factor & $0.1$ \\
      \# iterations for conjugate gradient & $10$ \\
      \# iterations for the value function & $5$ \\
      Minibatch size for the value function & $128$ \\
      \bottomrule
  \end{tabular}
\end{table}

\begin{table}[t!]
  \caption{Hyperparameters for A2C.}
  \label{table:hp_a2c}
  \centering
  \begin{tabular}{ll}
      \toprule
      Hyperparameter & Value \\
      \midrule
      Optimizer & RMSProp \\
      Adam learning rate & $1e-4$ \\
      Nonlinearity & ReLU \\
      \# decision steps per train step & $256$ \\
      \bottomrule
  \end{tabular}
\end{table}

\end{document}